\documentclass{article}

\usepackage[preprint, nonatbib]{neurips_2025}
\usepackage{subcaption}
\usepackage{graphicx}

\usepackage[utf8]{inputenc} 

\usepackage[backend=biber, maxnames=4, minnames=1, bibstyle=ieee, citestyle=numeric-comp, url=false, doi=true]{biblatex}
\AtEveryBibitem{\clearfield{issn}}

\usepackage[T1]{fontenc}    
\usepackage{hyperref}       
\usepackage{url}            
\usepackage{booktabs}       
\usepackage{amsfonts}       
\usepackage{nicefrac}       
\usepackage{microtype}      
\usepackage{xcolor}         
\usepackage{graphicx}

\usepackage[compact]{titlesec}
\titlespacing*{\section}{0pt}{*0.8}{*0.4}
\titlespacing*{\subsection}{0pt}{*0.7}{*0.3}
\titlespacing*{\subsubsection}{0pt}{*0.6}{*0.2}

\title{Generalization Beyond Benchmarks: \\ Evaluating Learnable Protein-Ligand Scoring Functions on Unseen Targets}

\author{
\centerline{Jakub Kopko$^{1}$\thanks{First author.} \qquad David Graber$^{2,3}$ \qquad Saltuk Mustafa Eyrilmez$^{4,5}$} \\
\centerline{\textbf{Stanislav Mazurenko}$^{4,5}$ \qquad \textbf{David Bednar}$^{4,5}$ \qquad \textbf{Jiri Sedlar}$^{1}$ \qquad \textbf{Josef Sivic}$^{1}$} \vspace{0.2cm} \\
$^{1}$CIIRC, Faculty of Electrical Engineering, Czech Technical University in Prague \\
$^{2}$Seminar for Applied Mathematics, Department of Mathematics, ETH Zurich \\
$^{3}$Institute for Computational Life Sciences, Zurich University of Applied Sciences \\
$^{4}$Loschmidt Laboratories, Faculty of Science, Masaryk University, Brno \\
$^{5}$ICRC, St.~Anne’s University Hospital, Brno \vspace{0.2cm} \\
\texttt{\{jakub.kopko, josef.sivic\}@cvut.cz} \\
\texttt{david.graber@sam.math.ethz.ch}
}

\begin{document}

\maketitle
\vspace{-0.5cm}

\begin{abstract}

As machine learning becomes increasingly central to molecular design, it is vital to ensure the reliability of learnable protein–ligand scoring functions on novel protein targets. While many scoring functions perform well on standard benchmarks, their ability to generalize beyond training data remains a significant challenge. In this work, we evaluate the generalization capability of state-of-the-art scoring functions on dataset splits that simulate evaluation on targets with a limited number of known structures and experimental affinity measurements. Our analysis reveals that the commonly used benchmarks do not reflect the true challenge of generalizing to novel targets. We also investigate whether large-scale self-supervised pretraining can bridge this generalization gap and we provide preliminary evidence of its potential. Furthermore, we probe the efficacy of simple methods that leverage limited test-target data to improve scoring function performance. Our findings underscore the need for more rigorous evaluation protocols and offer practical guidance for designing scoring functions with predictive power extending to novel protein targets.

\end{abstract}

\section{Introduction}

Structure-based drug discovery seeks to identify molecules that bind with high affinity and selectivity to a target protein based on structural information of the protein and ligand. In this setting, machine learning (ML) has been applied in two major directions. First, ML models are increasingly used to generate or predict energetically favorable protein-ligand conformations (binding poses)~\cite{Abramson2024, Krishna2024, Wohlwend2025, Corso2022}, complementing or replacing traditional docking algorithms. Second, ML supports the evaluation of these poses through learnable scoring functions, which estimate binding affinities and guide the ranking of candidate compounds~\cite{Stepniewska2018, Li2019, graber_resolving_2025, Shen2023}. Scoring functions play a central role in molecular docking, by evaluating predicted binding poses, and in virtual screening, by prioritizing compounds from large chemical libraries~\cite{WALTERS1998160}. Beyond structure-based approaches, ML is also applied in structure-free settings to screen compounds without requiring explicit protein–ligand complexes~\cite{lam2024protein-d8c}. In recent years, ML models have achieved impressive performance on standard benchmarks~\cite{Harren2024, Valsson2025}, driving growing interest in their integration into real-world drug discovery pipelines.

However, the ability of ML-based scoring functions to generalize to novel protein targets remains an important and increasingly relevant challenge, as many therapeutically significant proteins are poorly represented in current datasets. PLINDER~\cite{durairaj2024plinder-080}, a database that clusters protein--ligand complexes according to their structural similarity, provides a useful view of this problem. In the PDBbind database~\cite{Wang2004, Liu2015}, for example, some medically important targets, including the APOE4 variant~\cite{Hunsberger2019} and the KEAP1 Kelch domain~\cite{LI200454750}, have only a few available complexes. APOE4, a major genetic risk factor for Alzheimer’s disease~\cite{Hunsberger2019}, appears as a small cluster of only two complexes with nearly identical ligands. KEAP1, an important target in cancer research~\cite{Adinolfi2023}, is represented by just 18 highly similar pockets in PDBbind. Beyond naturally occurring proteins, advances in protein design, increasingly accelerated by machine learning~\cite{Notin2024}, are beginning to produce proteins with novel binding sites~\cite{Watson2023}. Designing such sites de novo offers advantages over repurposing natural proteins: it can enable functions not found in nature and allows integration of engineering principles such as tunability, controllability, and modularity directly into the scaffold. In synthetic biology, for example, de novo proteins are being developed to create new metabolic or signaling pathways, where small-molecule modulators could provide external control over these functions~\cite{Camacho2018}. Scoring models must therefore be able to evaluate potential small-molecule interactions in scenarios that lie far outside their training data distribution. As the prevalence of such cases grows, understanding model performance on structurally novel, low-data targets becomes essential.

\begin{figure*}[ht]
    \centering
    \begin{subfigure}{0.32\textwidth}
        \includegraphics[width=\linewidth]{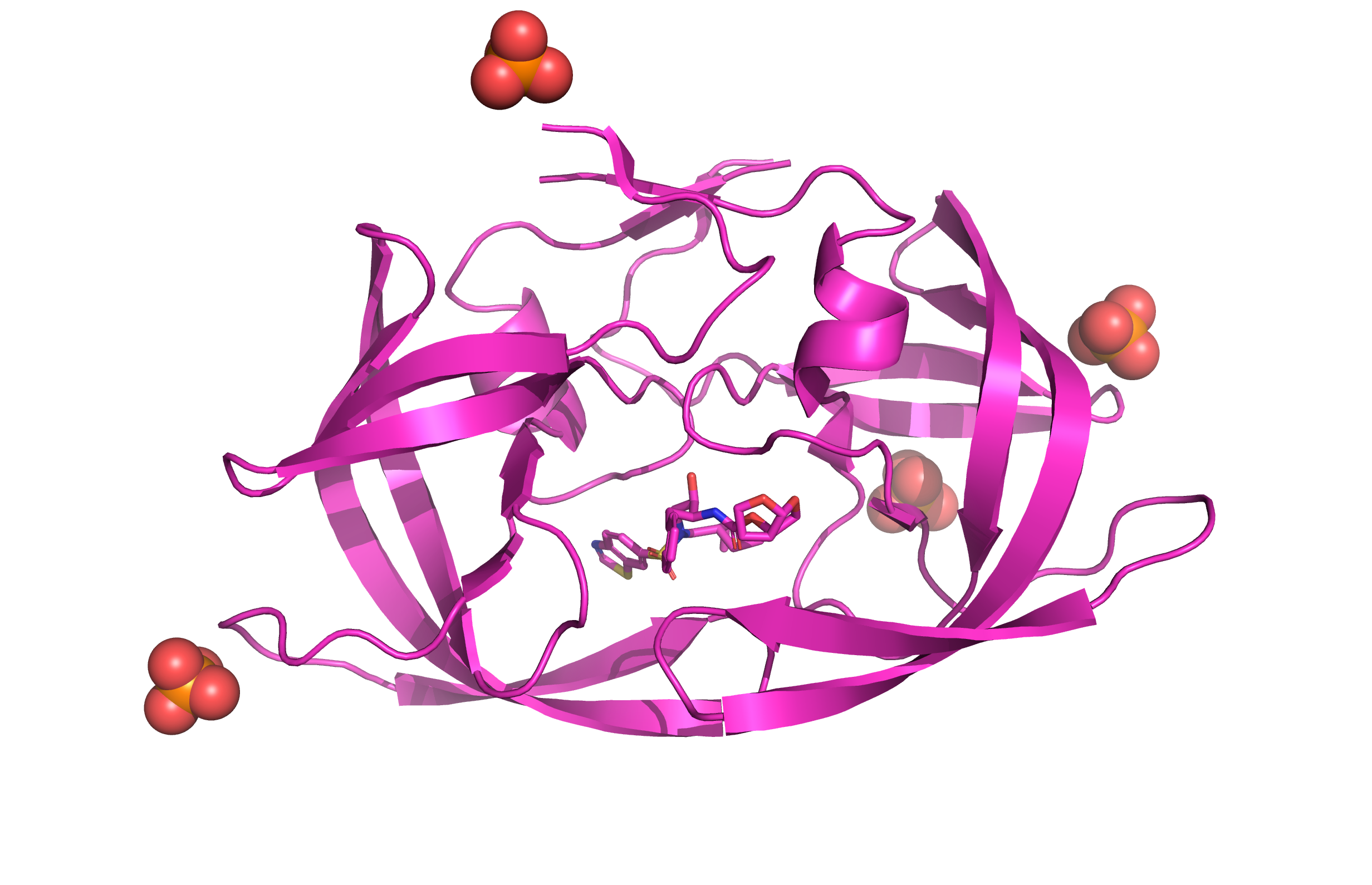}
    \end{subfigure}
    \hfill 
    \begin{subfigure}{0.32\textwidth}
        \includegraphics[width=\linewidth]{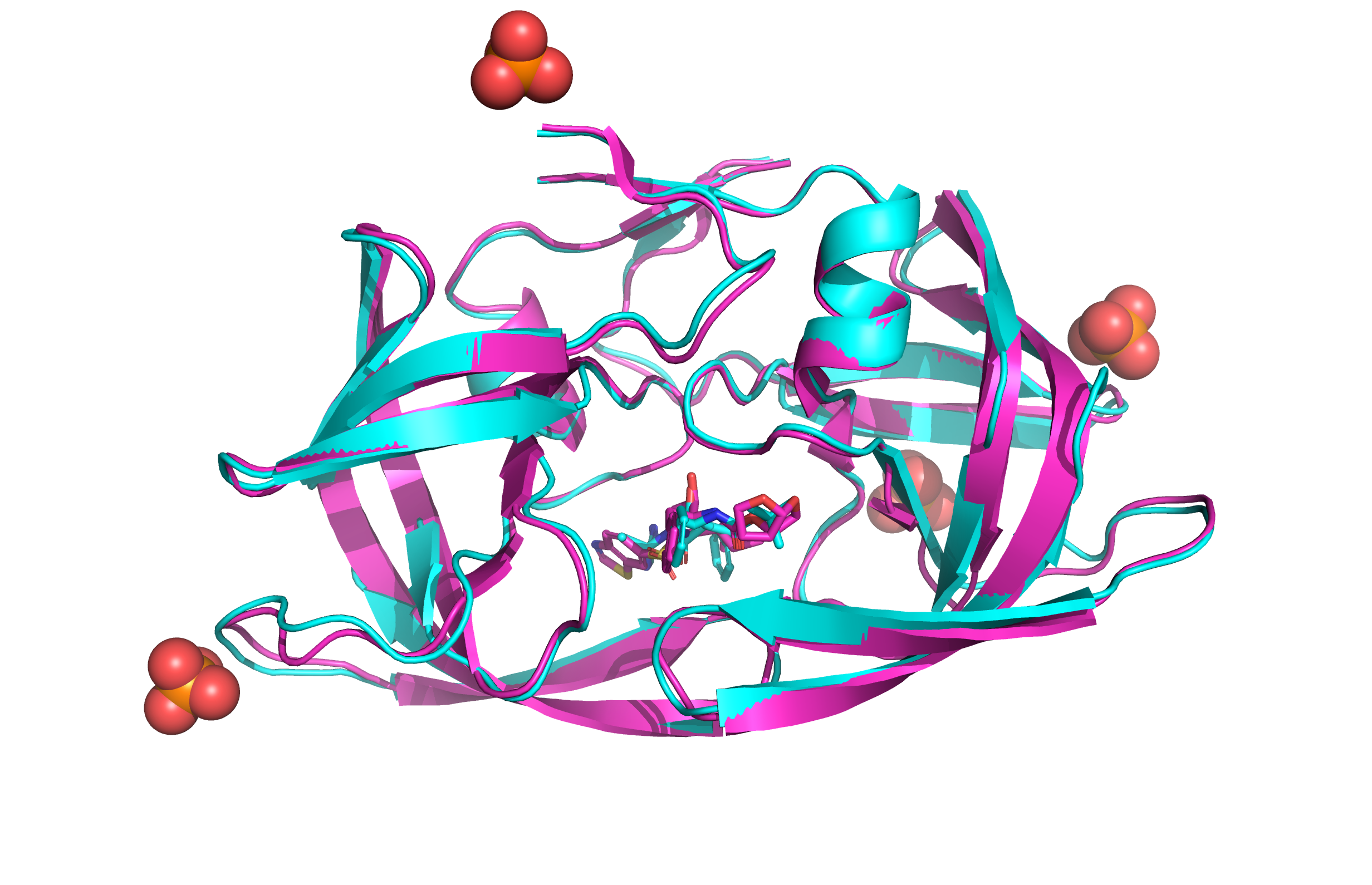}
    \end{subfigure}
    \hfill
    \begin{subfigure}{0.32\textwidth}
        \includegraphics[width=\linewidth]{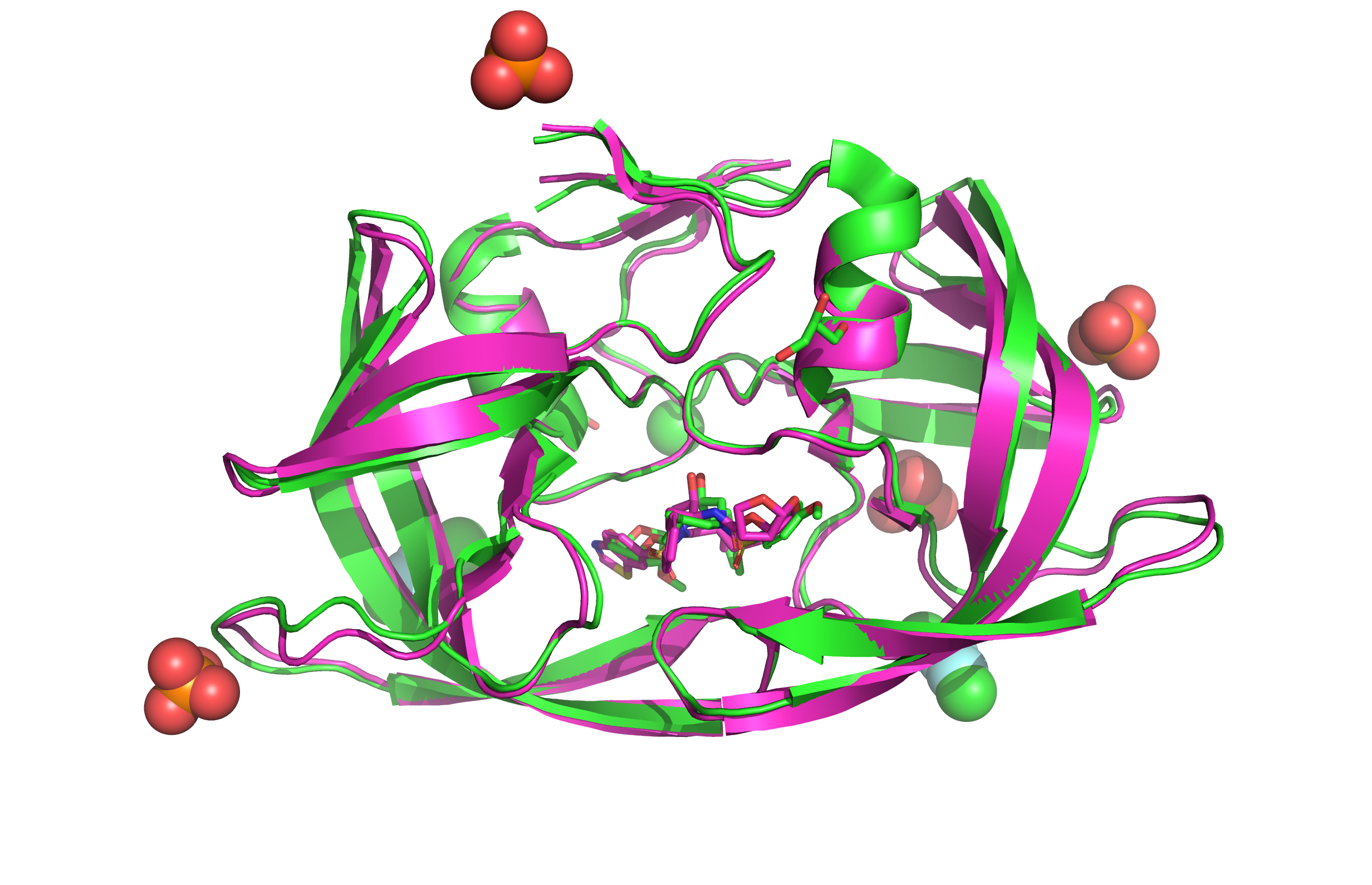}
    \end{subfigure}
    
    \vspace{2pt} 

    \begin{subfigure}{0.32\textwidth}
        \includegraphics[width=\linewidth]{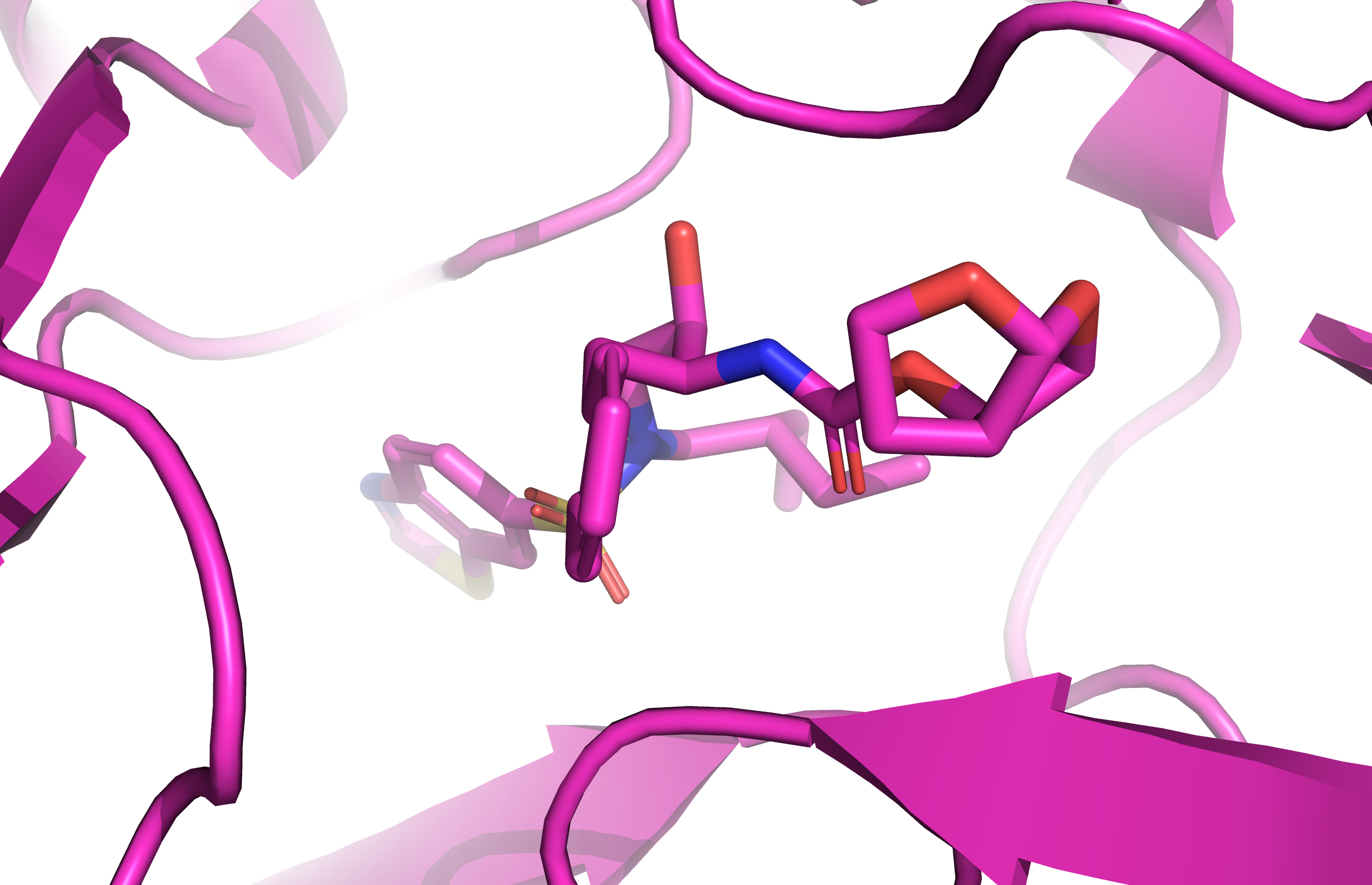}
    \end{subfigure}
    \hfill
    \begin{subfigure}{0.32\textwidth}
        \includegraphics[width=\linewidth]{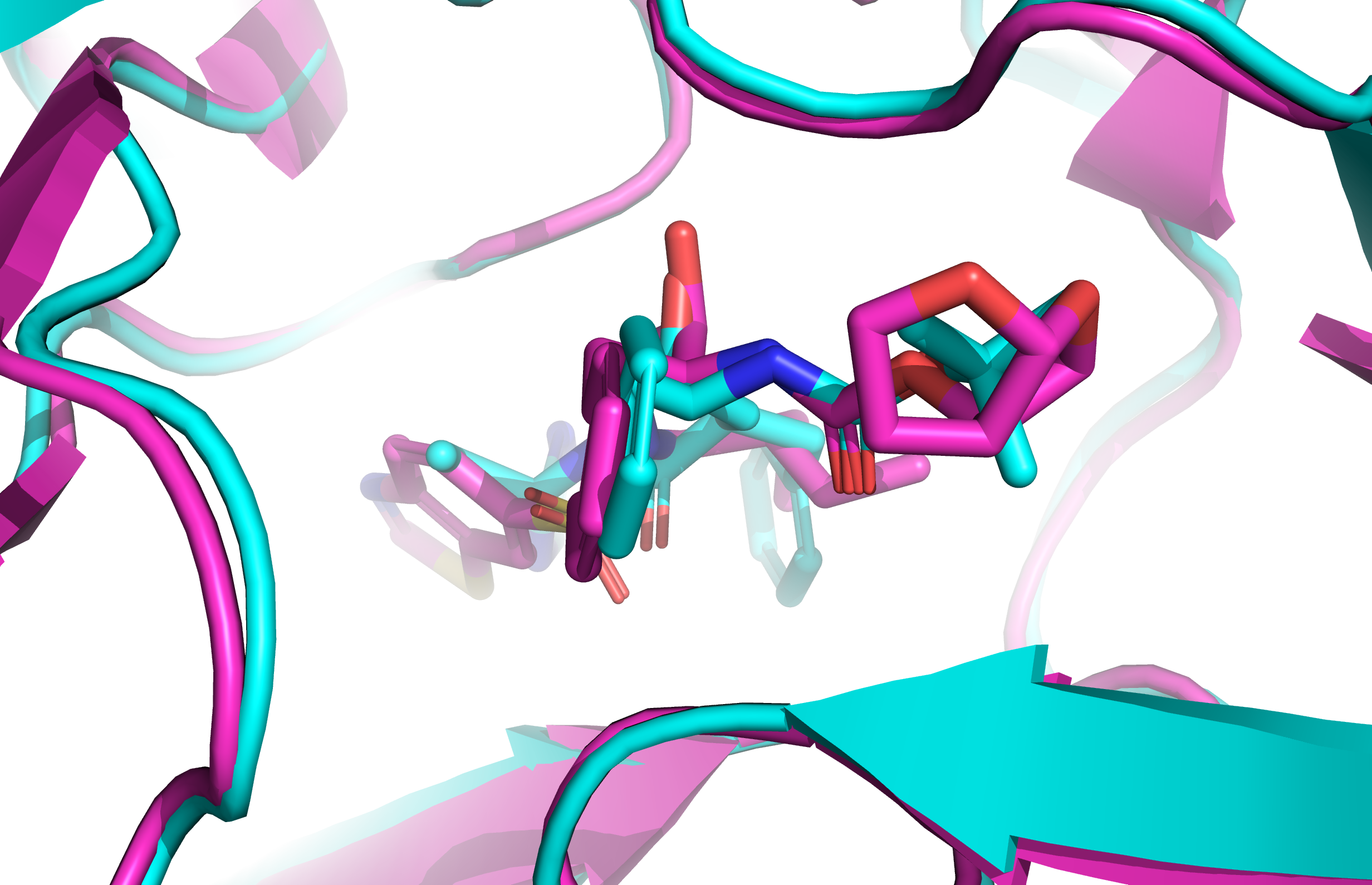}
    \end{subfigure}
    \hfill
    \begin{subfigure}{0.32\textwidth}
        \includegraphics[width=\linewidth]{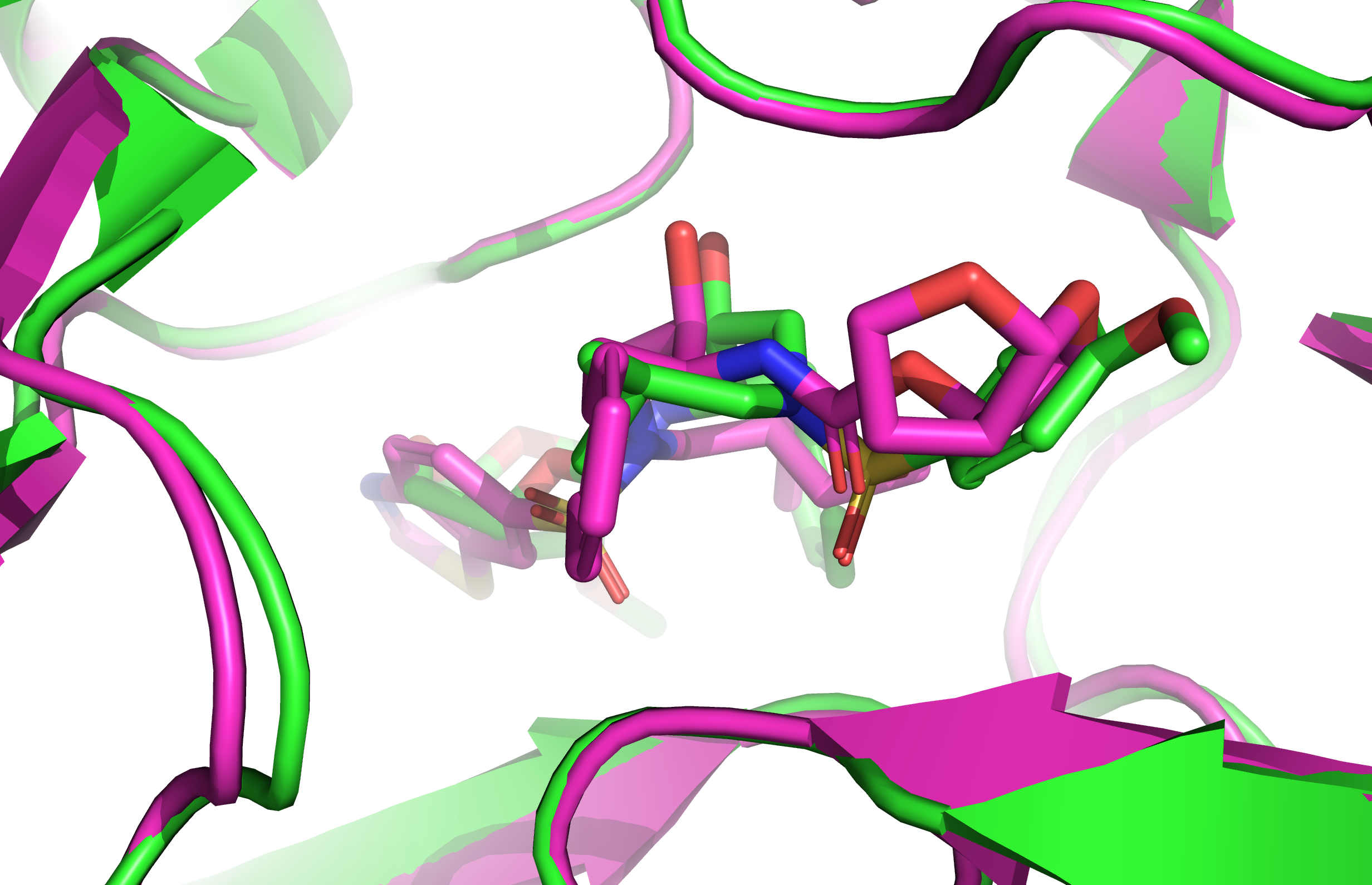}
    \end{subfigure}

    \caption{The full proteins (\textbf{top}) and their pockets (\textbf{bottom}) ... [rest of caption] ... }
    \label{fig:info_leakage}
\end{figure*}

Despite encouraging benchmark results, the ability of current models to generalize to unseen targets or unfamiliar chemical scaffolds remains uncertain~\cite{Yang2020, Volkov2022}. Several studies have revealed data leakage and inflated performance estimates~\cite{Kramer2010, Li2017, graber_resolving_2025, sellner2023quality-5ff}, primarily driven by biases in training datasets where protein families such as kinases and proteases are overrepresented and many complexes share high structural and ligand similarity. Widely used benchmarks, including CASF-2016~\cite{Su2019}, DUD-E~\cite{Mysinger2012}, and DEKOIS2.0~\cite{Bauer2013}, often contain target--ligand combinations resembling those in the training data (see Figure~\ref{fig:info_leakage}), which leads to overly optimistic performance estimates and obscures the true difficulty of generalizing to novel proteins. Many of those failures stem from dataset and evaluation biases rather than inherent model limitations. Overcoming these biases is essential for reliable deployment of ML-based scoring functions in drug discovery.

\paragraph{Contributions.} The contributions of this work are threefold. First, we present a systematic evaluation of two top-performing machine learning–based scoring functions: GEMS~\cite{graber_resolving_2025} and GenScore~\cite{Shen2023}. To rigorously assess their ability to generalize, we construct a strict series of dataset splits that limit pocket similarity between training and test sets, providing a framework for out-of-distribution (OOD) evaluation of scoring functions. In contrast to previous leakage analyses, which have mainly focused on docking and pose-generation tasks~\cite{Buttenschoen2024,corso2024deep,durairaj2024plinder-080}, our work targets the scoring problem directly and evaluates two models that have not been tested under such stringent conditions. Second, to explore potential solutions, we further investigate whether large-scale self-supervised pre-training with ATOMICA~\cite{fang2025atomica-617} embeddings can improve generalization by capturing atomic-scale interaction features across molecular modalities. These representations show preliminary promise in bridging the performance gap on novel targets. Finally, we investigate the scenario where a small number of experimental affinity measurements for the target ligand is available. We explore how to incorporate such measurements in validation or fine-tuning setups.   
Overall, our analysis highlights critical limitations in current evaluation practices and introduces new directions for developing more robust and generalizable scoring functions for real-world drug discovery applications.

\section{Related work}

\paragraph{Computational methods for protein-ligand scoring.}
Structure-based drug discovery (SBDD) uses three-dimensional structural information of biomolecules—typically proteins—to identify or develop new drugs. A central task in SBDD is identifying ligands that bind to a target protein with high affinity and specificity. A critical component of this process is scoring, which refers to estimating the strength of a protein–ligand interaction given a binding pose. Scoring enables two key applications: docking, where the goal is to identify the most likely binding pose of a ligand, and virtual screening, where compounds are ranked by predicted binding affinity to prioritize the most promising candidates.

Traditional physics-based scoring functions vary significantly in their computational cost and accuracy. Molecular mechanics (MM) methods (e.g., \cite{Trott2010,Friesner2004, Friesner2006}) are relatively efficient and capture basic physical interactions but they offer limited accuracy for binding affinity prediction. More sophisticated approaches exist, among which quantum mechanical (QM) methods (e.g., \cite{saltuk_rec, Pecina2024}) improve estimates by explicitly accounting for electronic effects. Free energy perturbation (FEP)~\cite{Cournia2021}, in contrast, is not merely a scoring function but a simulation-based method that can provide rigorous estimates of ligand binding free energies through extensive conformational sampling. However, both QM and FEP approaches are far too computationally expensive to be applied in large-scale virtual screening.

\paragraph{Machine learning for protein–ligand scoring.}
Recent machine learning (ML) models aim to offer a combination of speed and accuracy by learning scoring functions directly from structural data. Modern architectures such as graph neural networks~\cite{Karlov2020, Jiang2021, Li2023, Cao2024}, graph transformers~\cite{Shen2022, Shen2023}, and equivariant models that respect three-dimensional geometric symmetries~\cite{Cao2024, Bronstein2021} can model complex interaction patterns directly from raw atomic coordinates. Some models also explore pose-free scoring~\cite{lam2024protein-d8c}, i.e., predicting binding affinity directly from the protein sequence or structure and ligand representation (e.g., SMILES) without requiring a bound pose; however, these approaches fall outside the scope of this work.

Most ML-based models are trained on PDBbind crystal structures~\cite{Wang2004, Liu2015}, commonly using either experimental binding affinity labels or predicting protein–ligand interatomic distance distributions with a mixture density network (MDN)~\cite{Shen2023}, whose output probabilities are used to compute the final score. Another notable approach used denoising score matching for unsupervised binding energy prediction~\cite{jin2023dsmbind}. While ML scoring models have achieved strong benchmark performance, their ability to generalize to unseen targets remains an open question.

\paragraph{Information leakage in protein–ligand datasets.}

Many structure-based protein–ligand benchmarks suffer from ligand leakage, where structurally similar or identical binders appear in both training and test sets with similar affinity~\cite{Mastropietro2023}. Another issue is that binders and non-binders are drawn from different sources (e.g., ChEMBL vs.\ ZINC in DEKOIS~\cite{Bauer2013,brocidiacono2024improvedmetricbenchmarkassessing}), enabling models to exploit dataset-specific biases~\cite{Yang2020,Chen2019}. These artifacts affect benchmarks such as DUD-E~\cite{Mysinger2012} and DEKOIS2.0~\cite{Bauer2013}. While more recent efforts like LIT-PCBA~\cite{Tran-Nguyen2020} and BayesBind~\cite{brocidiacono2024improvedmetricbenchmarkassessing} attempt to mitigate such artifacts, they still face duplication and leakage issues~\cite{huang2025dataleakageredundancylitpcba}.

Leakage on the protein level is an equally important concern: high sequence or structural similarity between training and test proteins can produce overly optimistic performance estimates. Several approaches have been proposed to address this issue, evolving over time toward stricter and more targeted splits: PoseBusters~\cite{Buttenschoen2024} found that performance of many docking models degrades under stricter sequence-level splits; DockGen~\cite{corso2024deep} reduced redundancy using ECOD~\cite{ecod} domain splits; and PLINDER~\cite{durairaj2024plinder-080} provided clustering based on different metrics (pocket, ligand, or interaction similarity), allowing tailored splitting according to the task, and demonstrated that DiffDock~\cite{Corso2022} underperformed in these de-leaked settings. PDBbind CleanSplit, introduced alongside the simple GEMS model~\cite{graber_resolving_2025}, was designed to remove the most stringent similarities between PDBbind and the CASF benchmark. On this split, GEMS outperformed larger, more complex models, suggesting that some of the apparent progress on CASF may have resulted from the greater capacity of the complex models to memorize and overfit.

Our study concentrates on protein-side generalization to novel targets, emphasizing evaluation scenarios where the receptor itself is absent from training.

\paragraph{Large-scale self-supervised pretraining.}
Self-supervised pretraining on large unlabeled protein and molecular datasets has proven highly effective for downstream biochemical tasks. Protein language models such as ESM2 \cite{Lin2022} and ANKH \cite{Elnaggar2023} capture biophysical properties relevant for protein function, while chemical language models such as ChemBERTA~\cite{Ahmad2022} learn representations of small molecules. We adopt the GEMS scoring function~\cite{graber_resolving_2025}, which enhances graph neural networks with protein and chemical language model embeddings to improve binding affinity prediction.

Beyond sequence and chemical models, large-scale structural pretraining has also recently emerged; ATOMICA~\cite{fang2025atomica-617} is a geometric deep learning model trained on over two million interaction complexes using self-supervised denoising and masking. ATOMICA learns atomic-scale representations across proteins, small molecules, ions, lipids, and nucleic acids, yielding a compositional latent space that encodes shared physicochemical principles. In this work, we hypothesize and test whether the breadth of ATOMICA’s training enables better generalization in protein–ligand scoring tasks.

\begin{figure}[ht!]
    \centering
    \begin{tabular}{c c}
        \includegraphics[height=0.25\textheight]{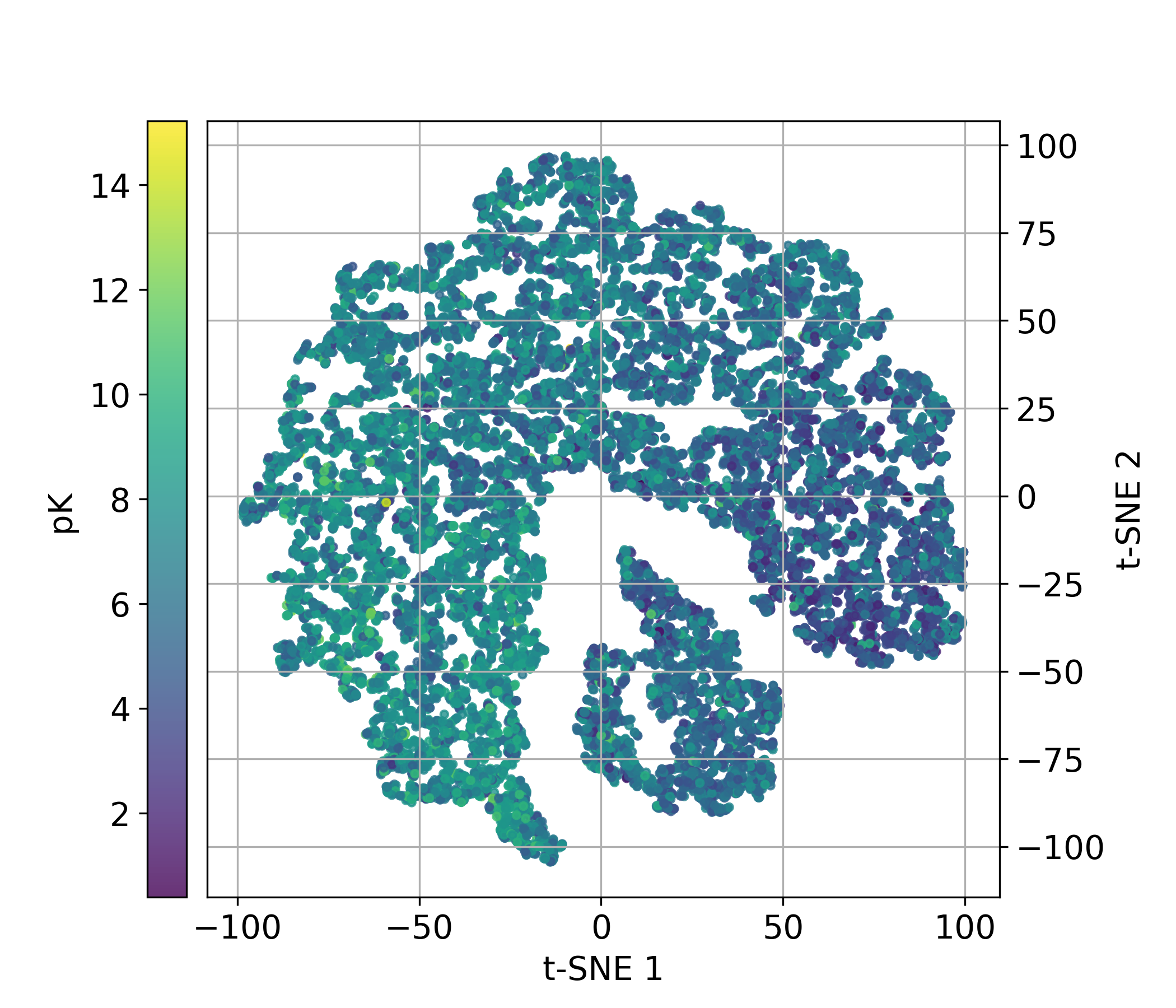} &
        \includegraphics[height=0.25\textheight]{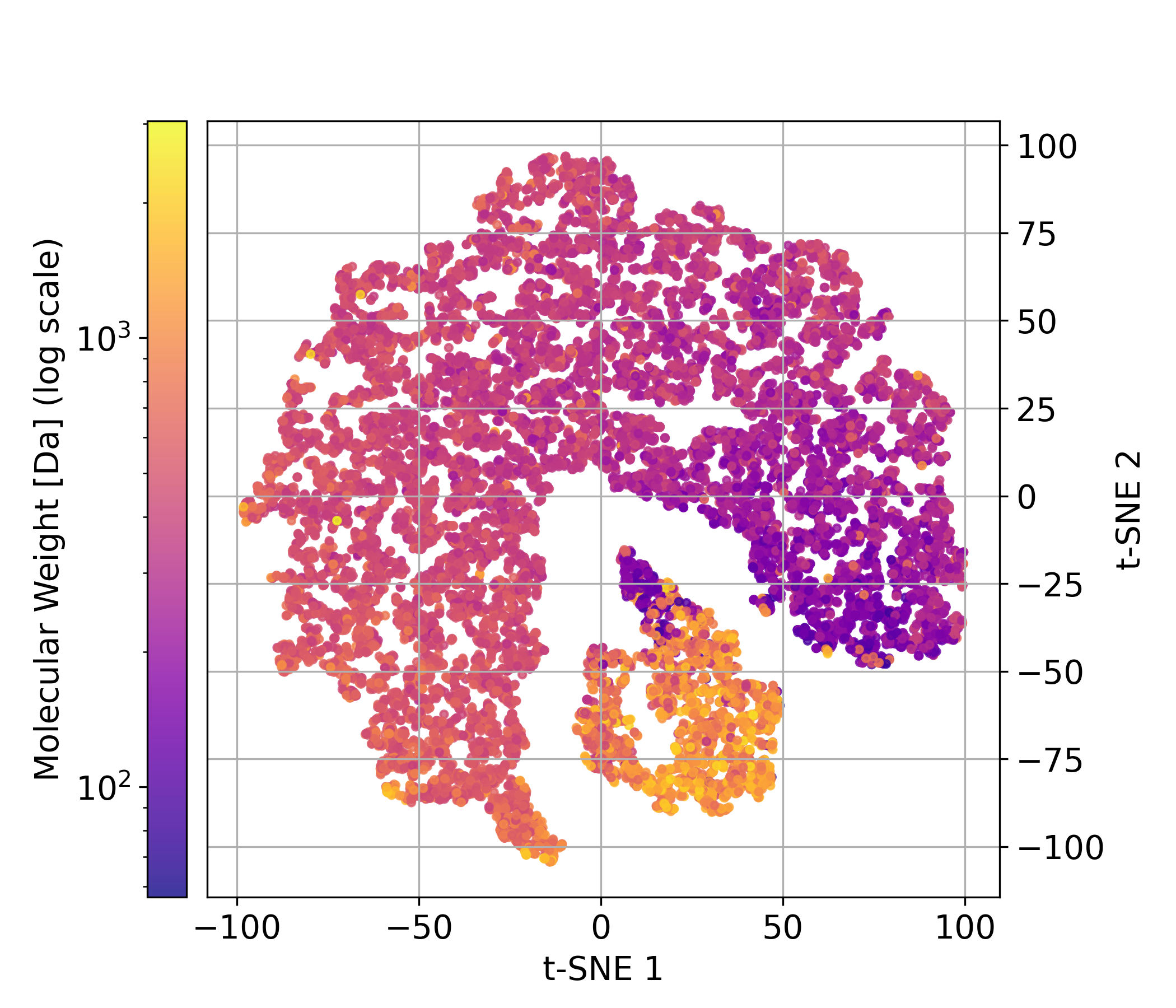} \\
    \end{tabular}
    \caption{%
    t-SNE projections of ATOMICA embeddings of ligand–pocket interactions from PDBbind,
    colored by experimental affinity (\textbf{left}) and molecular weight (\textbf{right}). The large, crescent-shaped cluster shows clear gradients of affinity and molecular weight, while the smaller, well-separated cluster contains most of the largest ligands, indicating that ATOMICA space captures both properties.
    }
    \label{fig:tsne_aff}
\end{figure}

\section{Experimental setup}

This section outlines the datasets, splitting strategies, training procedures, and evaluation metrics used in our study. We also describe methods for leveraging limited test-distribution data and for visualizing ATOMICA embeddings in two dimensions, providing intuition into the structure of the learned representation space.

\paragraph{Datasets.}
We use PDBbind (v.2020)~\cite{Wang2004, Liu2015} as the primary dataset for training, given its wide adoption and availability of experimental binding affinity labels (approximately 19{,}000 complexes). For benchmarking, we employ the well-established CASF-2016 dataset~\cite{Su2019}, comprising 285 high-quality complexes. Additionally, we use PLINDER~\cite{durairaj2024plinder-080} for clustering and splitting PDBbind complexes. Finally, we make use of ATOMICA~\cite{fang2025atomica-617} embeddings (\texttt{ATOMICA\_pretrained}), obtained via self-supervised pretraining with masking and denoising tasks on the Cambridge Structural Database (CSD)~\cite{Groom2016CSD}, which comprises roughly 1.75 million structures, and on Q-BioLiP~\cite{yang2013biolip-bfd,qbiolip}, with approximately 300{,}000 entries. While Q-BioLiP contains complexes overlapping with PDBbind, ATOMICA relies only on self-supervised tasks of coordinate denoising and masked block identity prediction, using information available at inference. We therefore do not consider this a source of data leakage.

\paragraph{Train–test splitting.}\label{para:splitting} 
We first evaluate models using the original PDBbind CleanSplit benchmark, testing on CASF targets as a baseline (Table~\ref{tab:scoring_cleansplit}). This establishes a reference point for performance when near-duplicate complexes are removed from the training dataset based on a joint assessment of ligand, pose, and pocket similarity.  

To construct stricter OOD evaluations, we further prepare our own train–test splits based on PLINDER’s pocket-level clustering using lDDT~\cite{durairaj2024plinder-080} (cluster type \texttt{pocket\_lddt\_\_50\_\_community}). Seven CASF targets were selected to represent diverse protein families (PDB IDs: 1NVQ, 1SQA, 2P15, 2VW5, 3DD0, 3F3E, 3O9I). For each target, all complexes from its cluster are assigned to a separate test set, while the remaining data is used for training and validation. This approach ensures that the test sets contain pockets that are structurally distinct from those seen during training, providing a more stringent evaluation of OOD generalization (see Figure~\ref{fig:info_leakage}).
We also apply CleanSplit filtering to the training and validation sets of our OOD splits, enabling evaluation on CASF without obvious leakage and enabling direct comparison between CASF and OOD performance (see Figure~\ref{fig:OOD_training}).
Throughout this work, we refer to clusters by the PDB ID of one of the complexes used to retrieve them (e.g., the ``1NVQ cluster''). Visual inspection suggests that clusters usually contain highly similar pockets, often preserving the protein fold. Nonetheless, the clustering procedure can, in principle, detect similar pockets arising in different folds.

\paragraph{Training.}
To create the training datasets, the preprocessed graph representations of PDBbind published alongside GEMS and GenScore were filtered based on the complex IDs. Models were then trained in a 5-fold cross-validation (CV) setting with label-based stratification. To enable direct comparisons, the original GEMS and GenScore implementations were trained on the same 5-fold CV splits, using their respective published training scripts and setups. To generate predictions on the test datasets, the five models obtained from cross-validation were combined into an ensemble via prediction averaging.

\paragraph{Evaluation metrics.}  
We focused on evaluating scoring power, quantified by the Pearson correlation between predicted and true affinities. The affinity labels are based on experimental measurements of $K_i$, $K_d$, or IC$_{50}$ and are included as p$K$ values (p$K$ = --log($K_i$/$K_d$/IC$_{50}$)). To further demonstrate that the observed performance drop is not merely a consequence of reduced training set size, we include figures illustrating performance evolution on both the CASF benchmark and our test sets (Figure~\ref{fig:OOD_training} and Supplementary Figure \ref{fig:OOD_training_suppl}). These figures clearly show that CASF performance, while slightly affected, remains very high and consistently above the performance on novel test targets.

Recognizing that a robust evaluation of scoring functions extends beyond scoring power, we assessed out-of-distribution docking and screening performance of GenScore on CASF-2016 targets from the 1NVQ cluster, using a model trained without 1NVQ cluster proteins. This cluster offers a relatively large sample of targets ($n=50$ for docking, $n=10$ for screening), whereas other clusters did not provide enough samples to yield stable, statistically significant results. As it represents only a single target, we report these results in Supplementary Figure~\ref{fig:OOD_VS_training}.

\paragraph{ATOMICA embeddings generation and projection.}\label{para:embeddings}  
To examine whether pretrained molecular representations capture properties of protein--ligand interactions that support better generalization of scoring models, we generated ATOMICA~\cite{fang2025atomica-617} graph-level embeddings for ligand--pocket complexes in PDBbind. Each embedding is a 32-dimensional vector, and we successfully obtained representations for 19,189 complexes (98.7\% of the dataset, see below).
For visualization, we employed t-distributed Stochastic Neighbor Embedding (t-SNE), a non-linear dimensionality reduction method that projects high-dimensional vectors into a low-dimensional space while preserving local similarity structure. We used two components and a perplexity of 30. This projection provides an intuitive view of how ligand--pocket complexes are arranged in the learned representation space, offering qualitative insight into the organization of ATOMICA embeddings through Figure~\ref{fig:tsne_aff} and Supplementary Figure~\ref{fig:tsne_cluster}.

\paragraph{Evaluated models.}
We evaluate the following four models.  
{\bf (1) GEMS}~\cite{graber_resolving_2025} is a graph neural network that integrates ligand embeddings from ChemBERTa~\cite{Ahmad2022} and protein embeddings from ESM2~\cite{Lin2022} and ANKH~\cite{Elnaggar2023}.  
{\bf (2) GenScore}~\cite{Shen2023} employs a graph transformer trained with a mixture density network (MDN) objective to model interatomic distance distributions, building upon the approach of RTMScore~\cite{Shen2022}.  
Next two methods use ATOMICA embeddings. In {\bf (3) \textbf{ATOMICA-MLP}}, we train a simple multilayer perceptron using ATOMICA embeddings. In {\bf  (4) \textbf{GEMS\textsubscript{ATOMICA}}}, we create a variant of the GEMS scoring function in which ChemBERTa ligand embeddings were concatenated with ATOMICA embeddings.  
Embedding extraction failed for 1.3\% of complexes, primarily due to ligand fragmentation issues. As a result, the training and validation sets for {GEMS\textsubscript{ATOMICA}} and {ATOMICA-MLP} were marginally smaller, reflecting the complexes for which embeddings could not be computed. The test sets reported in Tables~\ref{tab:scoring_cleansplit}, \ref{tab:scoring_ood}, and \ref{tab:scoring_tech} are identical to ensure a fair comparison across all evaluated models.

\paragraph{Leveraging limited data.}  
Laboratories often specialize in specific types of proteins and may obtain only a limited set of crystal structures with ligands and corresponding experimental affinity measurements for a protein of interest. Effectively leveraging such target-specific data is important to improve predictive models under realistic conditions.  
To study this, we hold out 25 complexes from each target-specific test set. These complexes are then used in two separate scenarios, with full consistency maintained between training, validation, and test splits, accounting for ATOMICA preprocessing errors:
\textbf{(i) Validation scenario}, where the 25 complexes serve solely as a fixed validation set, while the original training data remains unchanged. Models trained in this scenario are denoted as \textbf{$\mathrm{GEMS}^{\mathrm{VAL}}$} and \textbf{$\mathrm{GEMS}_{\mathrm{ATOMICA}}^{\mathrm{VAL}}$}.  
\textbf{(ii) Finetuning scenario}, where, starting from the model trained with the original stratified k-fold (SKF) procedure, the best model is further finetuned on the 25 held-out complexes using a small learning rate for 25 epochs. The original training and validation sets are kept unchanged to ensure comparability. These models are denoted as \textbf{$\mathrm{GEMS}^{\mathrm{FT-25}}$} and \textbf{$\mathrm{GEMS}_{\mathrm{ATOMICA}}^{\mathrm{FT-25}}$} for the ATOMICA variant.

\section{Results}

In this section, we present a comprehensive analysis of our results. We begin by visualizing ATOMICA embeddings with t-SNE, using experimental and structural labels to provide qualitative intuition about the organization of the representation space. We then evaluate model performance on the PDBbind CleanSplit benchmark, followed by assessment on our stricter out-of-distribution (OOD) splits. Finally, we investigate strategies for leveraging limited data from the test target distribution and compare their effectiveness.

\paragraph{PDBbind embeddings are well organized in ATOMICA space.}

We first inspect the projection of ATOMICA embeddings described in Section~\ref{para:embeddings}.  
In Figure~\ref{fig:tsne_aff}, we observe an affinity gradient along the larger, crescent-shaped cluster, which coincides with a gradient in molecular weight. Interestingly, the very heavy molecules appear to concentrate predominantly in the smaller, well-separated cluster.  
Supplementary Figure~\ref{fig:tsne_cluster} shows that the three clusters used for our OOD evaluation are also localized in distinct regions of the embedding space. Notably, the 3F3E cluster, which proved particularly challenging for GenScore (see Table~\ref{tab:scoring_ood}), is concentrated mostly in the smaller cluster corresponding to heavier molecules. 
Overall, our findings provide empirical and visual evidence confirming that the ATOMICA embedding space is well structured with respect to fundamental properties such as pocket shapes, ligand sizes, and binding affinity.

\paragraph{Performance on PDBbind CleanSplit.}

In Table~\ref{tab:scoring_cleansplit}, we evaluate the effect of enriching the GEMS scoring function with ATOMICA graph-level embeddings and compare it with the original GEMS and GenScore on the CASF benchmark after training on the full PDBbind CleanSplit. In this setup, we do not observe any advantage from incorporating the additional embeddings, with GEMS\textsubscript{ATOMICA} achieving marginally worse performance than the original GEMS (Table~\ref{tab:scoring_cleansplit}). Interestingly, while the simplistic ATOMICA-MLP is not competitive overall, it attains a reasonable performance on the independent subset of CASF (Table~\ref{tab:scoring_cleansplit}), highlighting that even simple models can capture some useful information from the embeddings.

\begin{table}[ht]
\centering
\small
\setlength{\tabcolsep}{6pt}
\caption{Scoring performance of different methods trained on PDBbind CleanSplit evaluated on CASF-2016 and the independent subset of CASF-2016~\cite{graber_resolving_2025}.}
\label{tab:scoring_cleansplit}
\begin{tabular}{lcc|cc}
\toprule
 & \multicolumn{2}{c|}{\textbf{Pearson correlation $\uparrow$}} & \multicolumn{2}{c}{\textbf{RMSE $\downarrow$}} \\
\cmidrule(lr){2-3} \cmidrule(lr){4-5}
 \textbf{Method} & \textbf{CASF-2016} & \textbf{CASF-2016 Indep.} & \textbf{CASF-2016} & \textbf{CASF-2016 Indep.} \\
\midrule
GenScore        & 0.807 & 0.737 & 1.280 & 1.580 \\
GEMS           & \textbf{0.815} & \textbf{0.828} & \textbf{1.274} & \textbf{1.347} \\
ATOMICA-MLP & 0.583 & 0.704 & 1.965 & 1.960 \\
GEMS\textsubscript{ATOMICA} & 0.808 & 0.822 & 1.292 & 1.361 \\
\bottomrule
\end{tabular}
\end{table}

\paragraph{Performance on strictly novel targets.}

We next evaluate model performance on the novel-target splits introduced in Section~\ref{para:splitting}. Scoring previously unseen proteins proves substantially more challenging than established benchmarks such as CASF. For example, the best Pearson correlation on a novel target is 0.736 (Table~\ref{tab:scoring_ood}, GEMS\textsubscript{ATOMICA}, target 1SQA), whereas the average across all clusters is only 0.470 for GEMS and 0.498 for GEMS\textsubscript{ATOMICA} (Table~\ref{tab:scoring_ood}). By comparison, both models achieve Pearson correlations of 0.815 and 0.808 on CASF, respectively (Table~\ref{tab:scoring_cleansplit}). Performance also varies widely across targets: while the 1SQA cluster yields relatively strong results, clusters such as 3DD0, 2P15, 3F3E, and 3O9I all result in correlations below 0.5 for every method, underscoring the difficulty imposed by strict generalization conditions.  

GEMS\textsubscript{ATOMICA} consistently outperforms the standard GEMS across all clusters except 1NVQ (Table~\ref{tab:scoring_ood}). It also achieves both the highest average and the highest minimum performance, indicating that incorporating ATOMICA embeddings improves the robustness of scoring protein–ligand interactions on unseen proteins. This demonstrates that more challenging benchmarks can reveal advantages of certain approaches that remain hidden on easier benchmarks like CASF.

There is no universal winner, however. GenScore achieves the best results on the 3DD0 and 3O9I clusters but fails completely on 3F3E, with a near-random correlation of 0.047 (Table~\ref{tab:scoring_ood}). Interestingly, even the simple ATOMICA-MLP model manages to reach top performance on the 2P15 cluster (Table~\ref{tab:scoring_ood}), showing that useful signal can be extracted from ATOMICA embeddings even with smaller modeling capacity.

\begin{table}[ht]
\centering
\small
\setlength{\tabcolsep}{5pt}
\caption{{\bf Performance on strictly novel targets.} Scoring power (Pearson correlation $r$ $\uparrow$) across different unseen protein clusters from the CASF benchmark. ``Avg.'' column reports the average across all protein clusters and ``Min.'' the minimum. }
\label{tab:scoring_ood}
\begin{tabular}{lccccccc|cc}
\toprule
\textbf{Method / Target} & \textbf{1NVQ} & \textbf{1SQA} & \textbf{2P15} & \textbf{2VW5} & \textbf{3DD0} & \textbf{3F3E} & \textbf{3O9I} & \textbf{Avg.} &
\textbf{Min.} \\
\midrule
\textit{\#Complexes}       & 2714 & 736 & 462 & 207 & 475 & 391 & 469 & --- & ---\\
\midrule
GenScore         & 0.451 & 0.696 & 0.415 & 0.384 & \textbf{0.481} & 0.047 & \textbf{0.480} & 0.422 & 0.047  \\
ATOMICA-MLP      & 0.540 & 0.631 & \textbf{0.479} & 0.524 & 0.346 & 0.149 & 0.400  & 0.438 & 0.149  \\
GEMS   & \textbf{0.609} & 0.717 & 0.364 & 0.551 & 0.236 & 0.404 & 0.410 & 0.470 & 0.236  \\
GEMS\textsubscript{ATOMICA} & 0.602 & \textbf{0.736} & 0.379 & \textbf{0.596} & 0.311 & \textbf{0.434} & 0.431 & \textbf{0.498} & \textbf{0.311}  \\
\bottomrule
\end{tabular}
\end{table}

\begin{figure}[ht!]
  \centering
  \includegraphics[width=\textwidth]{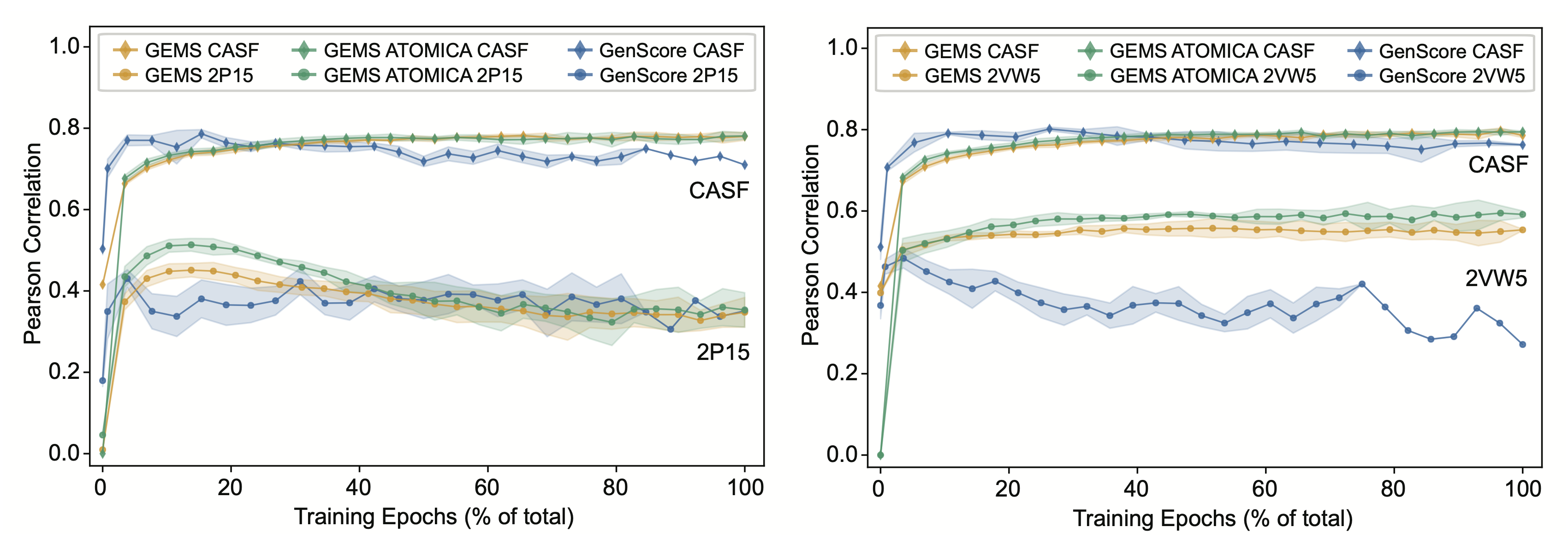}
  \caption{Evolution of the scoring power of the three scoring methods GenScore, GEMS and GEMS\textsubscript{ATOMICA} during model training with the original stratified k-fold splitting of the train–validation data, evaluated on CASF-2016 and the out-of-distribution (OOD) clusters 2P15 (left) and 2VW5 (right). The x-axis shows training progress as a percentage of total training epochs, while the y-axis displays the Pearson correlation coefficient $\uparrow$ between predicted and true affinity. Each line represents the mean performance across five cross-validation folds, with shaded uncertainty regions ($\pm1$ standard deviation) indicating the variability in performance across the five training runs. Note that the uncertainty regions disappear towards higher epoch numbers due to early stopping, which results in different models completing training at different epochs. Consequently, the later portions of the curves are based on fewer than five models. The final results imply that the original stratified k-fold splitting leads to overfitted models, and that using limited test-target validation data allows for better early stopping and the selection of better models.
  }
  \label{fig:OOD_training}
\end{figure}

\paragraph{Exploring strategies for utilizing limited data.}

We next examine strategies for leveraging a small number of additional target complexes, either by using them for validation or by fine-tuning on them (Table~\ref{tab:scoring_tech}). The results are mixed: both validation and fine-tuning improve performance on some clusters but degrade it on others. For instance, validation with ATOMICA embeddings improves performance on 2P15 and 3DD0, but decreases it on 3F3E. Similarly, fine-tuning boosts results on  3DD0 and 3O9I, yet reduces them on 1NVQ and 2VW5.

A motivating example for the validation strategy is shown in Figure~\ref{fig:OOD_training}. On the left, the 2P15 cluster illustrates that peak performance is reached early in training, visible as the hill-shaped curve of performance evolution. Without proper early stopping, all methods exhibit overfitting, and their performance declines to a similar level. Training dynamics varies across targets, as shown on the right side of the figure and in Supplementary Figure~\ref{fig:OOD_training_suppl}.

Despite these fluctuations, both strategies yield overall gains compared to training without access to the additional complexes (see GEMS and GEMS\textsubscript{ATOMICA} in Table~\ref{tab:scoring_ood}). In particular, they improve not only the average but also the worst-case (minimum) performance across clusters. Among them, the fine-tuned models achieve the strongest results, with $\mathrm{GEMS}_{\mathrm{ATOMICA}}^{\mathrm{FT-25}}$ delivering the highest average correlation (0.550) and $\mathrm{GEMS}^{\mathrm{FT-25}}$ attaining the highest minimum (0.386).  

\begin{table}[ht]
\centering
\small
\setlength{\tabcolsep}{5pt}
\caption{{\bf Performance on strictly novel targets with additional target-specific annotated data.} Scoring power (Pearson correlation $r$ $\uparrow$) across different test protein clusters of models utilizing the additional 25 target protein complexes for validation and fine-tuning.}
\label{tab:scoring_tech}
\begin{tabular}{lccccccc|cc}
\toprule
\textbf{Method / Target} & \textbf{1NVQ} & \textbf{1SQA} & \textbf{2P15} & \textbf{2VW5} & \textbf{3DD0} & \textbf{3F3E} & \textbf{3O9I} & \textbf{Avg.} & \textbf{Min.}  \\
\midrule
\textit{\#Complexes}       & 2689 & 711 & 437 & 182 & 450 & 366 & 444 & --- & --- \\
\midrule
\textbf{$\mathrm{GEMS}^{\mathrm{VAL}}$}   & 0.621 & 0.726 & 0.432 & 0.558 & 0.497 & 0.376 & 0.280 & 0.499 & 0.280  \\
\textbf{$\mathrm{GEMS}_{\mathrm{ATOMICA}}^{\mathrm{VAL}}$} & \textbf{0.626} & \textbf{0.742} & 0.514 & \textbf{0.602} & 0.419 & 0.375 & 0.354 & 0.519 & 0.375  \\
\midrule
\textbf{$\mathrm{GEMS}^{\mathrm{FT-25}}$}  & 0.563 & 0.670 & 0.556 & 0.537 & \textbf{0.620} & \textbf{0.386} & 0.424 & 0.537 & \textbf{0.386} \\
\textbf{$\mathrm{GEMS}_{\mathrm{ATOMICA}}^{\mathrm{FT-25}}$}  & 0.561 & 0.717 & \textbf{0.610} & 0.531 & 0.572 & 0.385 & \textbf{0.476} & \textbf{0.550} & 0.385 \\

\bottomrule
\end{tabular}
\end{table}

\section{Conclusions}

We studied the challenge of training scoring functions that generalize reliably to unseen proteins. Our results reveal a substantial gap between performance on standard benchmarks such as CASF and on stricter, novel-target splits. While CASF remains a useful reference, its optimistic estimates mask the difficulty of true generalization. Achieving robustness to entirely new proteins will likely require the models to capture the underlying physical principles of protein–ligand binding. These findings highlight the need for stricter, carefully designed benchmarks to become standard practice, particularly in scenarios where realistic performance estimates are essential.

Consistent with prior reports on GEMS~\cite{graber_resolving_2025}, our study suggests that recent gains in scoring may partly reflect improved data memorization rather than genuine generalization. Nevertheless, augmenting GEMS with ATOMICA embeddings consistently improved performance, showing that richer representations can enhance generalization. We also found that even limited target-specific data can narrow the generalization gap, boosting both average and worst-case performance.

\paragraph{Future directions.}  
Future work should emphasize more restrictive data splits and systematic reevaluation of existing methods to identify the drivers of true generalization, extending this analysis also to docking and screening scenarios, motivated by our preliminary results showing a performance drop on the 1NVQ cluster (see Supplementary Figure~\ref{fig:OOD_VS_training}). řExpanding our exploration of strategies for leveraging sparse target data, such as quantifying how performance gains scale with available data and testing more advanced fine-tuning techniques, represents another promising avenue. Finally, extending beyond graph-level ATOMICA embeddings to block-level (i.e., residue, nucleotide or chemical motif level) or atom-level representations may further strengthen protein–ligand modeling.

\newpage
\section*{Acknowledgements}

This work was supported by the EU's Horizon Europe Programme under the Grant agreement No.~101136607 (CLARA Project), and was co-funded by the EU from the Operational Programme Jan Amos Komenský (OP JAK) (project Center for Artificial Intelligence and Quantum Computing in System Brain Research, reg.~no.~CZ.02.01.01/00/23\_029/0008437). This work was further supported by the European Union (ERC FRONTIER No.~101097822, ELIAS No.~101120237) and by the Technology Agency of the Czech Republic under the NCC Programme (RETEMED TN02000122) and the NRP from the EU RRF (TEREP TN02000122/001N). The article reflects the author's view, and the Agency is not responsible for any use that may be made of the information it contains. Additional support was provided by the Czech Ministry of Education, Youth and Sports [ESFRI RECETOX RI LM2023069] and the Student Grant Competition of the Czech Technical University in Prague (SGS25/092/OHK3/2T/13). Computational resources were provided by e-INFRA CZ and ELIXIR-CZ [ID~90254, 90255, LM2023055]. We acknowledge VSB – Technical University of Ostrava, IT4Innovations National Supercomputing Center, Czech Republic, for awarding this project access to the LUMI supercomputer, owned by the EuroHPC Joint Undertaking, hosted by CSC (Finland) and the LUMI consortium through the Ministry of Education, Youth and Sports of the Czech Republic through the e-INFRA CZ (grant ID:~90254), through the 32nd Open Access Competition (project OPEN-32-3).

\newpage
\section*{References}
\printbibliography[heading=none]

@article{Bronstein2021, 
  year     = {2021}, 
  title    = {Geometric Deep Learning: Grids, Groups, Graphs, Geodesics, and Gauges}, 
  author   = {Bronstein, Michael M. and Bruna, Joan and Cohen, Taco and Veličković, Petar}, 
  url      = {http://arxiv.org/abs/2104.13478}
}

@article{Lin2022, 
  year     = {2022}, 
  title    = {Evolutionary-scale prediction of atomic level protein structure with a language model}, 
  author   = {Lin, Zeming and Akin, Halil and Rao, Roshan and Hie, Brian and Zhu, Zhongkai and Lu, Wenting and Smetanin, Nikita and Verkuil, Robert and Kabeli, Ori and Shmueli, Yaniv and Costa, Allan Dos Santos and Fazel-Zarandi, Maryam and Sercu, Tom and Candido, Salvatore and Rives, † Alexander}, 
  doi      = {10.1101/2022.07.20.500902}, 
  url      = {https://doi.org/10.1101/2022.07.20.500902}
}

@article{Elnaggar2023, 
  year     = {2023}, 
  title    = {Ankh: Optimized Protein Language Model Unlocks General-Purpose Modelling}, 
  author   = {Elnaggar, Ahmed and Essam, Hazem and Salah-Eldin, Wafaa and Moustafa, Walid and Elkerdawy, Mohamed and Rochereau, Charlotte and Rost, Burkhard}, 
  journal  = {{arXiv}}, 
  doi      = {10.48550/arxiv.2301.06568}, 
  eprint   = {2301.06568}
}

@article{Corso2022, 
  year     = {2022}, 
  title    = {{DiffDock}: Diffusion Steps, Twists, and Turns for Molecular Docking}, 
  author   = {Corso, Gabriele and Stärk, Hannes and Jing, Bowen and Barzilay, Regina and Jaakkola, Tommi}, 
  journal  = {{arXiv}}, 
  doi      = {10.48550/arxiv.2210.01776}, 
  eprint   = {2210.01776}
}

@article{yang2013biolip-bfd, 
  year     = {2013}, 
  title    = {{BioLiP}: a semi-manually curated database for biologically relevant ligand–protein interactions}, 
  author   = {Yang, Jianyi and Roy, Ambrish and Zhang, Yang}, 
  journal  = {Nucleic Acids Research}, 
  issn     = {0305-1048}, 
  doi      = {10.1093/nar/gks966}, 
  pmid     = {23087378}, 
  pmcid    = {{PMC}3531193}, 
  pages    = {D1096--D1103}, 
  number   = {D1}, 
  volume   = {41}
}

@article{Volkov2022, 
  year     = {2022}, 
  title    = {On the Frustration to Predict Binding Affinities from Protein–Ligand Structures with Deep Neural Networks}, 
  author   = {Volkov, Mikhail and Turk, Joseph-André and Drizard, Nicolas and Martin, Nicolas and Hoffmann, Brice and Gaston-Mathé, Yann and Rognan, Didier}, 
  journal  = {Journal of Medicinal Chemistry}, 
  issn     = {0022-2623}, 
  doi      = {10.1021/acs.jmedchem.2c00487}, 
  pmid     = {35608179}, 
  pages    = {7946--7958}, 
  number   = {11}, 
  volume   = {65}, 
  note     = {{PDBbind} 2019}
}

@article{Karlov2020, 
  year     = {2020}, 
  title    = {{graphDelta}: {MPNN} Scoring Function for the Affinity Prediction of Protein–Ligand Complexes}, 
  author   = {Karlov, Dmitry S. and Sosnin, Sergey and Fedorov, Maxim V. and Popov, Petr}, 
  journal  = {{ACS} Omega}, 
  issn     = {2470-1343}, 
  doi      = {10.1021/acsomega.9b04162}, 
  pmid     = {32201802}, 
  pmcid    = {{PMC}7081425}, 
  pages    = {5150--5159}, 
  number   = {10}, 
  volume   = {5}, 
  note     = {{PDBbind} 2018}
}

@article{Wang2004, 
  year     = {2004}, 
  title    = {The {PDBbind} Database: Collection of Binding Affinities for Protein-Ligand Complexes with Known Three-Dimensional Structures}, 
  author   = {Wang, Renxiao and Fang, Xueliang and Lu, Yipin and Wang, Shaomeng}, 
  journal  = {Journal of Medicinal Chemistry}, 
  issn     = {0022-2623}, 
  doi      = {10.1021/jm030580l}, 
  pmid     = {15163179}, 
  pages    = {2977--2980}, 
  number   = {12}, 
  volume   = {47}
}

@article{Li2023, 
  year     = {2023}, 
  title    = {{GIANT}: Protein-Ligand Binding Affinity Prediction via Geometry-aware Interactive Graph Neural Network}, 
  author   = {Li, Shuangli and Zhou, Jingbo and Xu, Tong and Huang, Liang and Wang, Fan and Xiong, Haoyi and Huang, Weili and Dou, Dejing and Xiong, Hui}, 
  journal  = {{IEEE} Transactions on Knowledge and Data Engineering}, 
  issn     = {1041-4347}, 
  doi      = {10.1109/tkde.2023.3314502}, 
  pages    = {1--17}, 
  number   = {99}, 
  volume   = {{PP}}, 
  note     = {{PDBbind} 2016 refined}
}

@article{Stepniewska2018, 
  year     = {2018}, 
  title    = {Development and evaluation of a deep learning model for protein–ligand binding affinity prediction}, 
  author   = {Stepniewska-Dziubinska, Marta M and Zielenkiewicz, Piotr and Siedlecki, Pawel}, 
  journal  = {Bioinformatics}, 
  issn     = {1367-4803}, 
  doi      = {10.1093/bioinformatics/bty374}, 
  pmid     = {29757353}, 
  pmcid    = {{PMC}6198856}, 
  pages    = {3666--3674}, 
  number   = {21}, 
  volume   = {34}, 
  note     = {{PDBbind} 2016}
}

@article{Yang2020, 
  year     = {2020}, 
  title    = {Predicting or Pretending: Artificial Intelligence for Protein-Ligand Interactions Lack of Sufficiently Large and Unbiased Datasets}, 
  author   = {Yang, Jincai and Shen, Cheng and Huang, Niu}, 
  journal  = {Frontiers in Pharmacology}, 
  issn     = {1663-9812}, 
  doi      = {10.3389/fphar.2020.00069}, 
  pmid     = {32161539}, 
  pmcid    = {{PMC}7052818}, 
  pages    = {69}, 
  volume   = {11}
}

@article{Jiang2021, 
  year     = {2021}, 
  title    = {{InteractionGraphNet}: A Novel and Efficient Deep Graph Representation Learning Framework for Accurate Protein–Ligand Interaction Predictions}, 
  author   = {Jiang, Dejun and Hsieh, Chang-Yu and Wu, Zhenxing and Kang, Yu and Wang, Jike and Wang, Ercheng and Liao, Ben and Shen, Chao and Xu, Lei and Wu, Jian and Cao, Dongsheng and Hou, Tingjun}, 
  journal  = {Journal of Medicinal Chemistry}, 
  issn     = {0022-2623}, 
  doi      = {10.1021/acs.jmedchem.1c01830}, 
  pmid     = {34878785}, 
  pages    = {18209--18232}, 
  number   = {24}, 
  volume   = {64}
}

@article{Li2019, 
  year     = {2019}, 
  title    = {{DeepAtom}: A Framework for Protein-Ligand Binding Affinity Prediction}, 
  author   = {Li, Yanjun and Rezaei, Mohammad A and Li, Chenglong and Li, Xiaolin and Wu, Dapeng}, 
  journal  = {{arXiv}}, 
  doi      = {10.48550/arxiv.1912.00318}, 
  eprint   = {1912.00318}
}

@article{Trott2010, 
  year     = {2010}, 
  title    = {{AutoDock} Vina: Improving the speed and accuracy of docking with a new scoring function, efficient optimization, and multithreading}, 
  author   = {Trott, Oleg and Olson, Arthur J.}, 
  journal  = {Journal of Computational Chemistry}, 
  issn     = {0192-8651}, 
  doi      = {10.1002/jcc.21334}, 
  pmid     = {19499576}, 
  pmcid    = {{PMC}3041641}, 
  pages    = {455--461}, 
  number   = {2}, 
  volume   = {31}
}

@article{Su2019, 
  year     = {2019}, 
  title    = {Comparative Assessment of Scoring Functions: The {CASF}-2016 Update}, 
  author   = {Su, Minyi and Yang, Qifan and Du, Yu and Feng, Guoqin and Liu, Zhihai and Li, Yan and Wang, Renxiao}, 
  journal  = {Journal of Chemical Information and Modeling}, 
  issn     = {1549-9596}, 
  doi      = {10.1021/acs.jcim.8b00545}, 
  pmid     = {30481020}, 
  pages    = {895--913}, 
  number   = {2}, 
  volume   = {59}
}

@article{Liu2015, 
  year     = {2015}, 
  title    = {{PDB}-wide collection of binding data: current status of the {PDBbind} database}, 
  author   = {Liu, Zhihai and Li, Yan and Han, Li and Li, Jie and Liu, Jie and Zhao, Zhixiong and Nie, Wei and Liu, Yuchen and Wang, Renxiao}, 
  journal  = {Bioinformatics}, 
  issn     = {1367-4803}, 
  doi      = {10.1093/bioinformatics/btu626}, 
  pmid     = {25301850}, 
  pages    = {405--412}, 
  number   = {3}, 
  volume   = {31}
}

@article{Ahmad2022, 
  year     = {2022}, 
  title    = {{ChemBERTa}-2: Towards Chemical Foundation Models}, 
  author   = {Ahmad, Walid and Simon, Elana and Chithrananda, Seyone and Grand, Gabriel and Ramsundar, Bharath}, 
  journal  = {{arXiv}}, 
  doi      = {10.48550/arxiv.2209.01712}, 
  eprint   = {2209.01712}
}

@article{Watson2023, 
  year     = {2023}, 
  title    = {De novo design of protein structure and function with {RFdiffusion}}, 
  author   = {Watson, Joseph L. and Juergens, David and Bennett, Nathaniel R. and Trippe, Brian L. and Yim, Jason and Eisenach, Helen E. and Ahern, Woody and Borst, Andrew J. and Ragotte, Robert J. and Milles, Lukas F. and Wicky, Basile I. M. and Hanikel, Nikita and Pellock, Samuel J. and Courbet, Alexis and Sheffler, William and Wang, Jue and Venkatesh, Preetham and Sappington, Isaac and Torres, Susana Vázquez and Lauko, Anna and Bortoli, Valentin De and Mathieu, Emile and Ovchinnikov, Sergey and Barzilay, Regina and Jaakkola, Tommi S. and {DiMaio}, Frank and Baek, Minkyung and Baker, David}, 
  journal  = {Nature}, 
  issn     = {0028-0836}, 
  doi      = {10.1038/s41586-023-06415-8}, 
  pmid     = {37433327}, 
  pmcid    = {{PMC}10468394}, 
  pages    = {1089--1100}, 
  number   = {7976}, 
  volume   = {620}
}

@article{sellner2023quality-5ff, 
  year     = {2023}, 
  title    = {Quality Matters: Deep Learning-Based Analysis of Protein-Ligand Interactions with Focus on Avoiding Bias}, 
  author   = {Sellner, Manuel S. and Lill, Markus A. and Smieško, Martin}, 
  journal  = {{bioRxiv}}, 
  doi      = {10.1101/2023.11.13.566916}, 
  pages    = {2023.11.13.566916}
}

@article{Krishna2024, 
  year     = {2024}, 
  title    = {Generalized biomolecular modeling and design with {RoseTTAFold} All-Atom}, 
  author   = {Krishna, Rohith and Wang, Jue and Ahern, Woody and Sturmfels, Pascal and Venkatesh, Preetham and Kalvet, Indrek and Lee, Gyu Rie and Morey-Burrows, Felix S. and Anishchenko, Ivan and Humphreys, Ian R. and {McHugh}, Ryan and Vafeados, Dionne and Li, Xinting and Sutherland, George A. and Hitchcock, Andrew and Hunter, C. Neil and Kang, Alex and Brackenbrough, Evans and Bera, Asim K. and Baek, Minkyung and {DiMaio}, Frank and Baker, David}, 
  journal  = {Science}, 
  issn     = {0036-8075}, 
  doi      = {10.1126/science.adl2528}, 
  pmid     = {38452047}, 
  pages    = {eadl2528}, 
  number   = {6693}, 
  volume   = {384}
}

@article{Harren2024, 
  year     = {2024}, 
  title    = {Modern machine‐learning for binding affinity estimation of protein–ligand complexes: Progress, opportunities, and challenges}, 
  author   = {Harren, Tobias and Gutermuth, Torben and Grebner, Christoph and Hessler, Gerhard and Rarey, Matthias}, 
  journal  = {Wiley Interdisciplinary Reviews: Computational Molecular Science}, 
  issn     = {1759-0876}, 
  doi      = {10.1002/wcms.1716}, 
  number   = {3}, 
  volume   = {14}
}

@article{Li2017, 
  year     = {2017}, 
  title    = {Structural and Sequence Similarity Makes a Significant Impact on Machine-Learning-Based Scoring Functions for Protein–Ligand Interactions}, 
  author   = {Li, Yang and Yang, Jianyi}, 
  journal  = {Journal of Chemical Information and Modeling}, 
  issn     = {1549-9596}, 
  doi      = {10.1021/acs.jcim.7b00049}, 
  pmid     = {28358210}, 
  pages    = {1007--1012}, 
  number   = {4}, 
  volume   = {57}
}

@article{Kramer2010, 
  year     = {2010}, 
  title    = {Leave-Cluster-Out Cross-Validation Is Appropriate for Scoring Functions Derived from Diverse Protein Data Sets}, 
  author   = {Kramer, Christian and Gedeck, Peter}, 
  journal  = {Journal of Chemical Information and Modeling}, 
  issn     = {1549-9596}, 
  doi      = {10.1021/ci100264e}, 
  pmid     = {20936880}, 
  pages    = {1961--1969}, 
  number   = {11}, 
  volume   = {50}
}

@article{durairaj2024plinder-080, 
  year     = {2024}, 
  title    = {{PLINDER}: The protein-ligand interactions dataset and evaluation resource}, 
  author   = {Durairaj, Janani and Adeshina, Yusuf and Cao, Zhonglin and Zhang, Xuejin and Oleinikovas, Vladas and Duignan, Thomas and {McClure}, Zachary and Robin, Xavier and Studer, Gabriel and Kovtun, Daniel and Rossi, Emanuele and Zhou, Guoqing and Veccham, Srimukh and Isert, Clemens and Peng, Yuxing and Sundareson, Prabindh and Akdel, Mehmet and Corso, Gabriele and Stärk, Hannes and Tauriello, Gerardo and Carpenter, Zachary and Bronstein, Michael and Kucukbenli, Emine and Schwede, Torsten and Naef, Luca}, 
  journal  = {{bioRxiv}}, 
  doi      = {10.1101/2024.07.17.603955}, 
  pages    = {2024.07.17.603955}
}

@article{Abramson2024, 
  year     = {2024}, 
  title    = {Accurate structure prediction of biomolecular interactions with {AlphaFold} 3}, 
  author   = {Abramson, Josh and Adler, Jonas and Dunger, Jack and Evans, Richard and Green, Tim and Pritzel, Alexander and Ronneberger, Olaf and Willmore, Lindsay and Ballard, Andrew J. and Bambrick, Joshua and Bodenstein, Sebastian W. and Evans, David A. and Hung, Chia-Chun and O’Neill, Michael and Reiman, David and Tunyasuvunakool, Kathryn and Wu, Zachary and Žemgulytė, Akvilė and Arvaniti, Eirini and Beattie, Charles and Bertolli, Ottavia and Bridgland, Alex and Cherepanov, Alexey and Congreve, Miles and Cowen-Rivers, Alexander I. and Cowie, Andrew and Figurnov, Michael and Fuchs, Fabian B. and Gladman, Hannah and Jain, Rishub and Khan, Yousuf A. and Low, Caroline M. R. and Perlin, Kuba and Potapenko, Anna and Savy, Pascal and Singh, Sukhdeep and Stecula, Adrian and Thillaisundaram, Ashok and Tong, Catherine and Yakneen, Sergei and Zhong, Ellen D. and Zielinski, Michal and Žídek, Augustin and Bapst, Victor and Kohli, Pushmeet and Jaderberg, Max and Hassabis, Demis and Jumper, John M.}, 
  journal  = {Nature}, 
  issn     = {0028-0836}, 
  doi      = {10.1038/s41586-024-07487-w}, 
  pmid     = {38718835}, 
  pmcid    = {{PMC}11168924}, 
  pages    = {493--500}, 
  number   = {8016}, 
  volume   = {630}
}

@article{Mastropietro2023, 
  year     = {2023}, 
  title    = {Learning characteristics of graph neural networks predicting protein–ligand affinities}, 
  author   = {Mastropietro, Andrea and Pasculli, Giuseppe and Bajorath, Jürgen}, 
  journal  = {Nature Machine Intelligence}, 
  doi      = {10.1038/s42256-023-00756-9}, 
  pages    = {1427--1436}, 
  number   = {12}, 
  volume   = {5}
}

@article{lam2024protein-d8c, 
  year     = {2024}, 
  title    = {Protein language models are performant in structure-free virtual screening}, 
  author   = {Lam, Hilbert Yuen In and Guan, Jia Sheng and Ong, Xing Er and Pincket, Robbe and Mu, Yuguang}, 
  journal  = {Briefings in Bioinformatics}, 
  issn     = {1467-5463}, 
  doi      = {10.1093/bib/bbae480}, 
  pmid     = {39327890}, 
  pmcid    = {{PMC}11427677}, 
  pages    = {bbae480}, 
  number   = {6}, 
  volume   = {25}
}

@article{Valsson2025, 
  year     = {2025}, 
  title    = {Narrowing the gap between machine learning scoring functions and free energy perturbation using augmented data}, 
  author   = {Valsson, Ísak and Warren, Matthew T. and Deane, Charlotte M. and Magarkar, Aniket and Morris, Garrett M. and Biggin, Philip C.}, 
  journal  = {Communications Chemistry}, 
  doi      = {10.1038/s42004-025-01428-y}, 
  pmid     = {39922899}, 
  pmcid    = {{PMC}11807228}, 
  pages    = {41}, 
  number   = {1}, 
  volume   = {8}
}

@article{Cao2024, 
  year     = {2024}, 
  title    = {Generic protein–ligand interaction scoring by integrating physical prior knowledge and data augmentation modelling}, 
  author   = {Cao, Duanhua and Chen, Geng and Jiang, Jiaxin and Yu, Jie and Zhang, Runze and Chen, Mingan and Zhang, Wei and Chen, Lifan and Zhong, Feisheng and Zhang, Yingying and Lu, Chenghao and Li, Xutong and Luo, Xiaomin and Zhang, Sulin and Zheng, Mingyue}, 
  journal  = {Nature Machine Intelligence}, 
  doi      = {10.1038/s42256-024-00849-z}, 
  pages    = {688--700}, 
  number   = {6}, 
  volume   = {6}, 
  note     = {{EquiScore}}
}

@article{Shen2022, 
  year     = {2022}, 
  title    = {Boosting Protein–Ligand Binding Pose Prediction and Virtual Screening Based on Residue–Atom Distance Likelihood Potential and Graph Transformer}, 
  author   = {Shen, Chao and Zhang, Xujun and Deng, Yafeng and Gao, Junbo and Wang, Dong and Xu, Lei and Pan, Peichen and Hou, Tingjun and Kang, Yu}, 
  journal  = {Journal of Medicinal Chemistry}, 
  issn     = {0022-2623}, 
  doi      = {10.1021/acs.jmedchem.2c00991}, 
  pmid     = {35917397}, 
  pages    = {10691--10706}, 
  number   = {15}, 
  volume   = {65}, 
  note     = {{RTMScore}}
}

@article{Friesner2004, 
  year     = {2004}, 
  title    = {Glide: A New Approach for Rapid, Accurate Docking and Scoring. 1. Method and Assessment of Docking Accuracy}, 
  author   = {Friesner, Richard A. and Banks, Jay L. and Murphy, Robert B. and Halgren, Thomas A. and Klicic, Jasna J. and Mainz, Daniel T. and Repasky, Matthew P. and Knoll, Eric H. and Shelley, Mee and Perry, Jason K. and Shaw, David E. and Francis, Perry and Shenkin, Peter S.}, 
  journal  = {Journal of Medicinal Chemistry}, 
  issn     = {0022-2623}, 
  doi      = {10.1021/jm0306430}, 
  pmid     = {15027865}, 
  pages    = {1739--1749}, 
  number   = {7}, 
  volume   = {47}
}

@article{Friesner2006, 
  year     = {2006}, 
  title    = {Extra Precision Glide: Docking and Scoring Incorporating a Model of Hydrophobic Enclosure for Protein-Ligand Complexes}, 
  author   = {Friesner, Richard A. and Murphy, Robert B. and Repasky, Matthew P. and Frye, Leah L. and Greenwood, Jeremy R. and Halgren, Thomas A. and Sanschagrin, Paul C. and Mainz, Daniel T.}, 
  journal  = {Journal of Medicinal Chemistry}, 
  issn     = {0022-2623}, 
  doi      = {10.1021/jm051256o}, 
  pmid     = {17034125}, 
  pages    = {6177--6196}, 
  number   = {21}, 
  volume   = {49}
}

@article{Shen2023, 
  year     = {2023}, 
  title    = {A generalized protein–ligand scoring framework with balanced scoring, docking, ranking and screening powers}, 
  author   = {Shen, Chao and Zhang, Xujun and Hsieh, Chang-Yu and Deng, Yafeng and Wang, Dong and Xu, Lei and Wu, Jian and Li, Dan and Kang, Yu and Hou, Tingjun and Pan, Peichen}, 
  journal  = {Chemical Science}, 
  issn     = {2041-6520}, 
  doi      = {10.1039/d3sc02044d}, 
  pmid     = {37538816}, 
  pmcid    = {{PMC}10395315}, 
  pages    = {8129--8146}, 
  number   = {30}, 
  volume   = {14}, 
  note     = {{GenScore}}
}

@article{Wohlwend2025, 
  year     = {2025}, 
  title    = {Boltz-1 Democratizing Biomolecular Interaction Modeling}, 
  author   = {Wohlwend, Jeremy and Corso, Gabriele and Passaro, Saro and Getz, Noah and Reveiz, Mateo and Leidal, Ken and Swiderski, Wojtek and Atkinson, Liam and Portnoi, Tally and Chinn, Itamar and Silterra, Jacob and Jaakkola, Tommi and Barzilay, Regina}, 
  journal  = {{bioRxiv}}, 
  doi      = {10.1101/2024.11.19.624167}, 
  pmid     = {39605745}, 
  pmcid    = {{PMC}11601547}, 
  pages    = {2024.11.19.624167}
}

@article{fang2025atomica-617, 
  year     = {2025}, 
  title    = {{ATOMICA}: Learning Universal Representations of Intermolecular Interactions}, 
  author   = {Fang, Ada and Zhang, Zaixi and Zhou, Andrew and Zitnik, Marinka}, 
  journal  = {{bioRxiv}}, 
  doi      = {10.1101/2025.04.02.646906}, 
  pmid     = {40291688}, 
  pmcid    = {{PMC}12026499}, 
  pages    = {2025.04.02.646906}
}

@article{Groom2016CSD,
  author    = {Groom, C. R. and Bruno, I. J. and Lightfoot, M. P. and Ward, S. C.},
  title     = {The Cambridge Structural Database},
  journal   = {Acta Crystallographica Section B: Structural Science, Crystal Engineering and Materials},
  year      = {2016},
  volume    = {72},
  pages     = {171--179},
  doi       = {10.1107/S2052520616003954}
}

@article{qbiolip,
    author = {Wei, Hong and Wang, Wenkai and Peng, Zhenling and Yang, Jianyi},
    title = {Q-BioLiP: A Comprehensive Resource for Quaternary Structure-based Protein–ligand Interactions},
    journal = {Genomics, Proteomics \& Bioinformatics},
    volume = {22},
    number = {1},
    pages = {qzae001},
    year = {2024},
    month = {01},
    issn = {1672-0229},
    doi = {10.1093/gpbjnl/qzae001},
    url = {https://doi.org/10.1093/gpbjnl/qzae001},
    eprint = {https://academic.oup.com/gpb/article-pdf/22/1/qzae001/58457634/qzae001.pdf},
}

@article{WALTERS1998160,
title = {Virtual screening—an overview},
journal = {Drug Discovery Today},
volume = {3},
number = {4},
pages = {160-178},
year = {1998},
issn = {1359-6446},
doi = {https://doi.org/10.1016/S1359-6446(97)01163-X},
url = {https://www.sciencedirect.com/science/article/pii/S135964469701163X},
author = {W.Patrick Walters and Matthew T Stahl and Mark A Murcko}
}

@article{Hunsberger2019,
  author    = {Hunsberger, H. C. and Pinky, P. D. and Smith, W. and Suppiramaniam, V. and Reed, M. N.},
  title     = {The role of APOE4 in Alzheimer's disease: strategies for future therapeutic interventions},
  journal   = {Neuronal Signal},
  year      = {2019},
  volume    = {3},
  number    = {2},
  pages     = {NS20180203},
  month     = {06},
  doi       = {10.1042/NS20180203},
  pmid      = {32269835},
  pmcid     = {PMC7104324},
  note      = {Epub 2019 Apr 18}
}

@article{LI200454750,
title = {Crystal Structure of the Kelch Domain of Human Keap1*},
journal = {Journal of Biological Chemistry},
volume = {279},
number = {52},
pages = {54750-54758},
year = {2004},
issn = {0021-9258},
doi = {https://doi.org/10.1074/jbc.M410073200},
url = {https://www.sciencedirect.com/science/article/pii/S0021925819632424},
author = {Xuchu Li and Donna Zhang and Mark Hannink and Lesa J. Beamer}
}

@article{Adinolfi2023,
  author    = {Adinolfi, S. and Patinen, T. and Jawahar Deen, A. and Pitkänen, S. and Härkönen, J. and Kansanen, E. and Küblbeck, J. and Levonen, A. L.},
  title     = {The KEAP1-NRF2 pathway: Targets for therapy and role in cancer},
  journal   = {Redox Biology},
  year      = {2023},
  volume    = {63},
  pages     = {102726},
  month     = {07},
  doi       = {10.1016/j.redox.2023.102726},
  pmid      = {37146513},
  pmcid     = {PMC10189287},
  note      = {Epub 2023 Apr 29}
}

@article{Notin2024,
  author    = {Notin, P. and Rollins, N. and Gal, Y. and Sander, C. and Marks, D.},
  title     = {Machine learning for functional protein design},
  journal   = {Nature Biotechnology},
  year      = {2024},
  volume    = {42},
  number    = {2},
  pages     = {216--228},
  doi       = {10.1038/s41587-024-02127-0},
  url       = {https://doi.org/10.1038/s41587-024-02127-0}
}

@article{Pecina2024,
  author    = {Pecina, Adam and Fanfrlík, Jindřich and Lepšík, Martin and Řezáč, Jan},
  title     = {SQM2.20: Semiempirical quantum-mechanical scoring function yields DFT-quality protein–ligand binding affinity predictions in minutes},
  journal   = {Nature Communications},
  year      = {2024},
  volume    = {15},
  number    = {1},
  pages     = {1127},
  doi       = {10.1038/s41467-024-45431-8},
  url       = {https://doi.org/10.1038/s41467-024-45431-8}
}

@incollection{Cournia2021,
  author    = {Cournia, Zoe and Chipot, Christophe and Roux, Benoît and York, Darrin M. and Sherman, Woody},
  title     = {Free Energy Methods in Drug Discovery—Introduction},
  booktitle = {Free Energy Methods in Drug Discovery: Current State and Future Directions},
  series    = {ACS Symposium Series},
  volume    = {1397},
  pages     = {1--38},
  publisher = {American Chemical Society},
  year      = {2021},
  doi       = {10.1021/bk-2021-1397.ch001},
  url       = {https://doi.org/10.1021/bk-2021-1397.ch001},
  isbn      = {9780841298064}
}

@article{Mysinger2012,
  title={Directory of Useful Decoys, Enhanced (DUD-E): Better Ligands and Decoys for Better Benchmarking},
  author={Mysinger, Michael M. and Carchia, Michael and Irwin, John J. and Shoichet, Brian K.},
  journal={Journal of Medicinal Chemistry},
  volume={55},
  number={14},
  pages={6582--6594},
  year={2012},
  publisher={American Chemical Society},
  doi={10.1021/jm300687e},
  url={https://doi.org/10.1021/jm300687e}
}

@article{Bauer2013,
  title={Evaluation and Optimization of Virtual Screening Workflows with DEKOIS 2.0 – A Public Library of Challenging Docking Benchmark Sets},
  author={Bauer, Matthias R. and Ibrahim, Tamer M. and Vogel, Simon M. and Boeckler, Frank M.},
  journal={Journal of Chemical Information and Modeling},
  volume={53},
  number={6},
  pages={1447--1462},
  year={2013},
  publisher={American Chemical Society},
  doi={10.1021/ci400115b},
  url={https://doi.org/10.1021/ci400115b}
}

@misc{brocidiacono2024improvedmetricbenchmarkassessing,
      title={An Improved Metric and Benchmark for Assessing the Performance of Virtual Screening Models}, 
      author={Michael Brocidiacono and Konstantin I. Popov and Alexander Tropsha},
      year={2024},
      eprint={2403.10478},
      archivePrefix={arXiv},
      primaryClass={q-bio.QM},
      url={https://arxiv.org/abs/2403.10478}, 
}

@article{Chen2019,
  title={Hidden bias in the DUD-E dataset leads to misleading performance of deep learning in structure-based virtual screening},
  author={Chen, L. and Cruz, A. and Ramsey, S. and Dickson, C. J. and Duca, J. S. and Hornak, V. and et al.},
  journal={PLoS ONE},
  volume={14},
  number={8},
  pages={e0220113},
  year={2019},
  doi={10.1371/journal.pone.0220113},
  url={https://doi.org/10.1371/journal.pone.0220113}
}

@article{Tran-Nguyen2020,
  title={LIT-PCBA: An Unbiased Data Set for Machine Learning and Virtual Screening},
  author={Tran-Nguyen, Viet-Khoa and Jacquemard, Célien and Rognan, Didier},
  journal={Journal of Chemical Information and Modeling},
  volume={60},
  number={9},
  pages={4263--4273},
  year={2020},
  doi={10.1021/acs.jcim.0c00155},
  url={https://doi.org/10.1021/acs.jcim.0c00155}
}

@misc{huang2025dataleakageredundancylitpcba,
      title={Data Leakage and Redundancy in the LIT-PCBA Benchmark}, 
      author={Amber Huang and Ian Scott Knight and Slava Naprienko},
      year={2025},
      eprint={2507.21404},
      archivePrefix={arXiv},
      primaryClass={cs.LG},
      url={https://arxiv.org/abs/2507.21404}, 
}

@article{Buttenschoen2024, title = {{{PoseBusters}}: {{AI-based}} Docking Methods Fail to Generate Physically Valid Poses or Generalise to Novel Sequences}, shorttitle = {{{PoseBusters}}}, author = {Buttenschoen, Martin and M  ~Deane, Charlotte}, year = {2024}, journal = {Chemical Science}, volume = {15}, number = {9}, pages = {3130--3139}, publisher = {Royal Society of Chemistry}, doi = {10.1039/D3SC04185A}, urldate = {2025-08-28}, langid = {english} }

@inproceedings{
corso2024deep,
title={Deep Confident Steps to New Pockets: Strategies for Docking Generalization},
author={Gabriele Corso and Arthur Deng and Nicholas Polizzi and Regina Barzilay and Tommi S. Jaakkola},
booktitle={The Twelfth International Conference on Learning Representations},
year={2024},
url={https://openreview.net/forum?id=UfBIxpTK10}
}

@article{ecod,
  title = {ECOD: an evolutionary classification of protein domains},
  author = {Cheng, Hao and Schaeffer, Robert D. and Liao, Yuanfang and Kinch, Lisa N. and Pei, Jimin and Shi, Sen and Kim, Byung H. and Grishin, Nick V.},
  journal = {PLoS Computational Biology},
  year = {2014},
  volume = {10},
  number = {12},
  pages = {e1003926},
  doi = {10.1371/journal.pcbi.1003926},
  url = {https://doi.org/10.1371/journal.pcbi.1003926}
}

@inproceedings{
jin2023dsmbind,
title={{DSMB}ind: an unsupervised generative modeling framework for binding energy prediction},
author={Wengong Jin and Caroline Uhler and Nir Hacohen},
booktitle={NeurIPS 2023 Generative AI and Biology (GenBio) Workshop},
year={2023},
url={https://openreview.net/forum?id=itp5RDMlNV}
}

@article{Camacho2018,
  title        = {Next-Generation Machine Learning for Biological Networks},
  author       = {Camacho, Diogo M. and Collins, Katherine M. and Powers, Rani K. and Costello, James C. and Collins, James J.},
  journal      = {Cell},
  year         = {2018},
  volume       = {173},
  number       = {7},
  pages        = {1581--1592},
  doi          = {10.1016/j.cell.2018.05.015},
  url          = {https://doi.org/10.1016/j.cell.2018.05.015},
  publisher    = {Elsevier},
  issn         = {0092-8674}
}

@article{saltuk_rec,
author = {Ryde, Ulf and S{\"o}derhjelm, Pär},
title = {Ligand-Binding Affinity Estimates Supported by Quantum-Mechanical Methods},
journal = {Chemical Reviews},
volume = {116},
number = {9},
pages = {5520-5566},
year = {2016},
doi = {10.1021/acs.chemrev.5b00630},
    note ={PMID: 27077817},

URL = { 
    
        https://doi.org/10.1021/acs.chemrev.5b00630
    
    

},
eprint = { 
    
        https://doi.org/10.1021/acs.chemrev.5b00630
    

}

}

@article{graber_resolving_2025,
	title = {Resolving data bias improves generalization in binding affinity prediction},
	volume = {7},
	copyright = {2025 The Author(s)},
	issn = {2522-5839},
	url = {https://www.nature.com/articles/s42256-025-01124-5},
	doi = {10.1038/s42256-025-01124-5},
	language = {en},
	number = {10},
	urldate = {2025-10-28},
	journal = {Nature Machine Intelligence},
	author = {Graber, David and Stockinger, Peter and Meyer, Fabian and Mishra, Siddhartha and Horn, Claus and Buller, Rebecca},
	month = oct,
	year = {2025},
	note = {Publisher: Nature Publishing Group},
	pages = {1713--1725},
}
\renewcommand{\thefigure}{S\arabic{figure}}
\renewcommand{\theHfigure}{S\arabic{figure}}
\setcounter{figure}{0}
\renewcommand{\figurename}{Supplementary Figure}
\newpage
\section*{Supplementary material}

\begin{figure}[ht!]
    \centering
    \begin{tabular}{c c}
        \includegraphics[height=0.25\textheight]{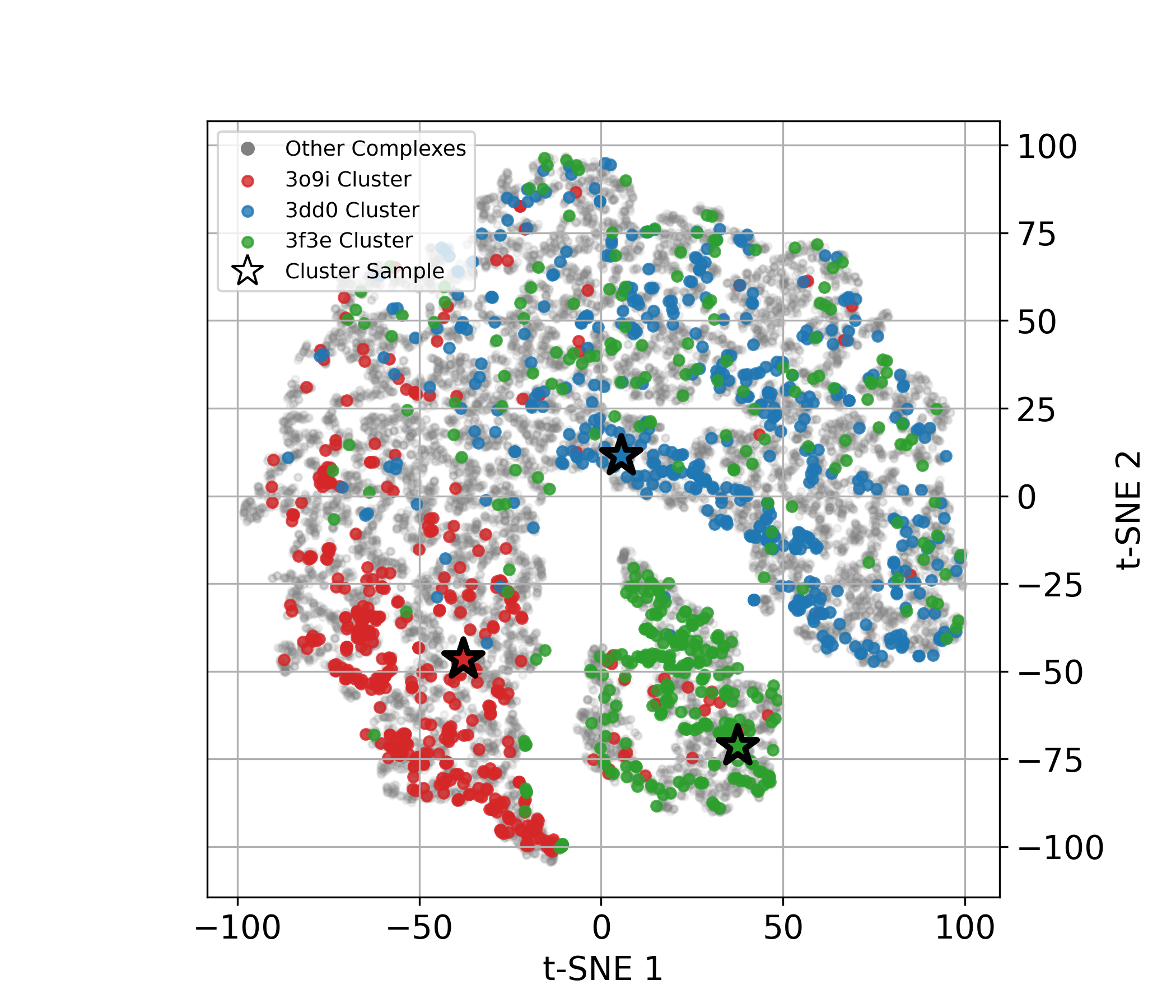} &
        \includegraphics[height=0.25\textheight]{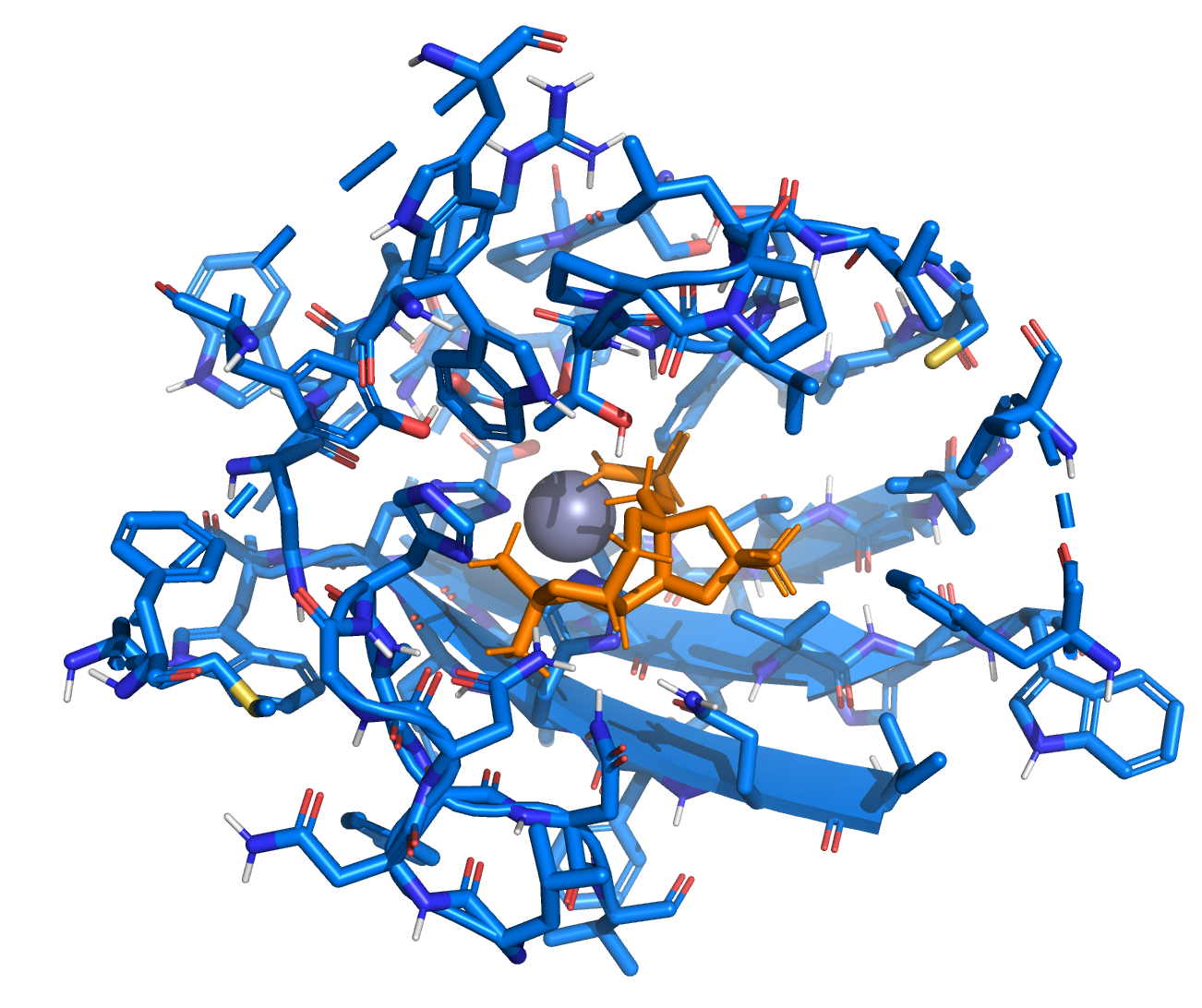} \\
        \includegraphics[height=0.25\textheight]{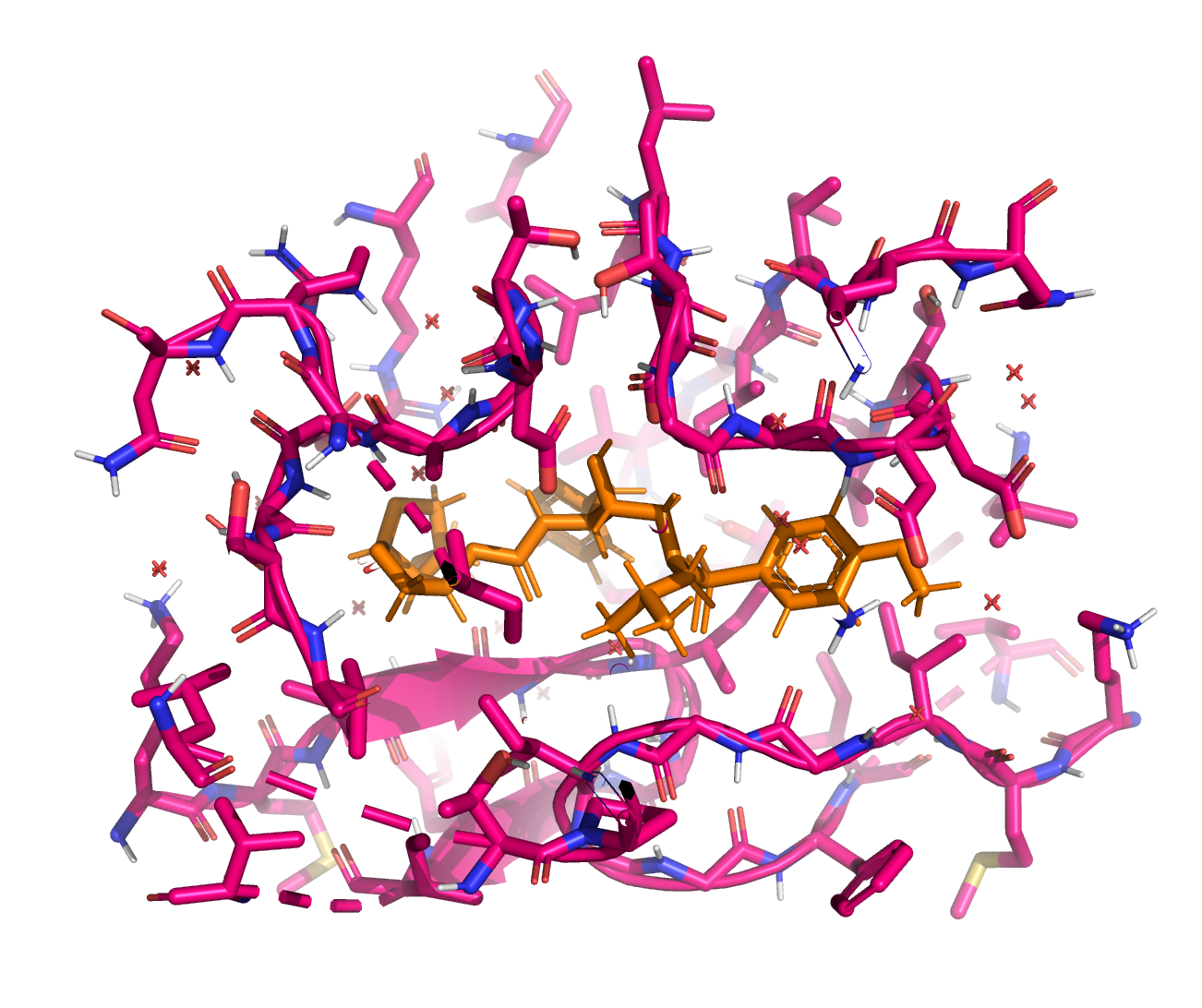} &
        \includegraphics[height=0.25\textheight]{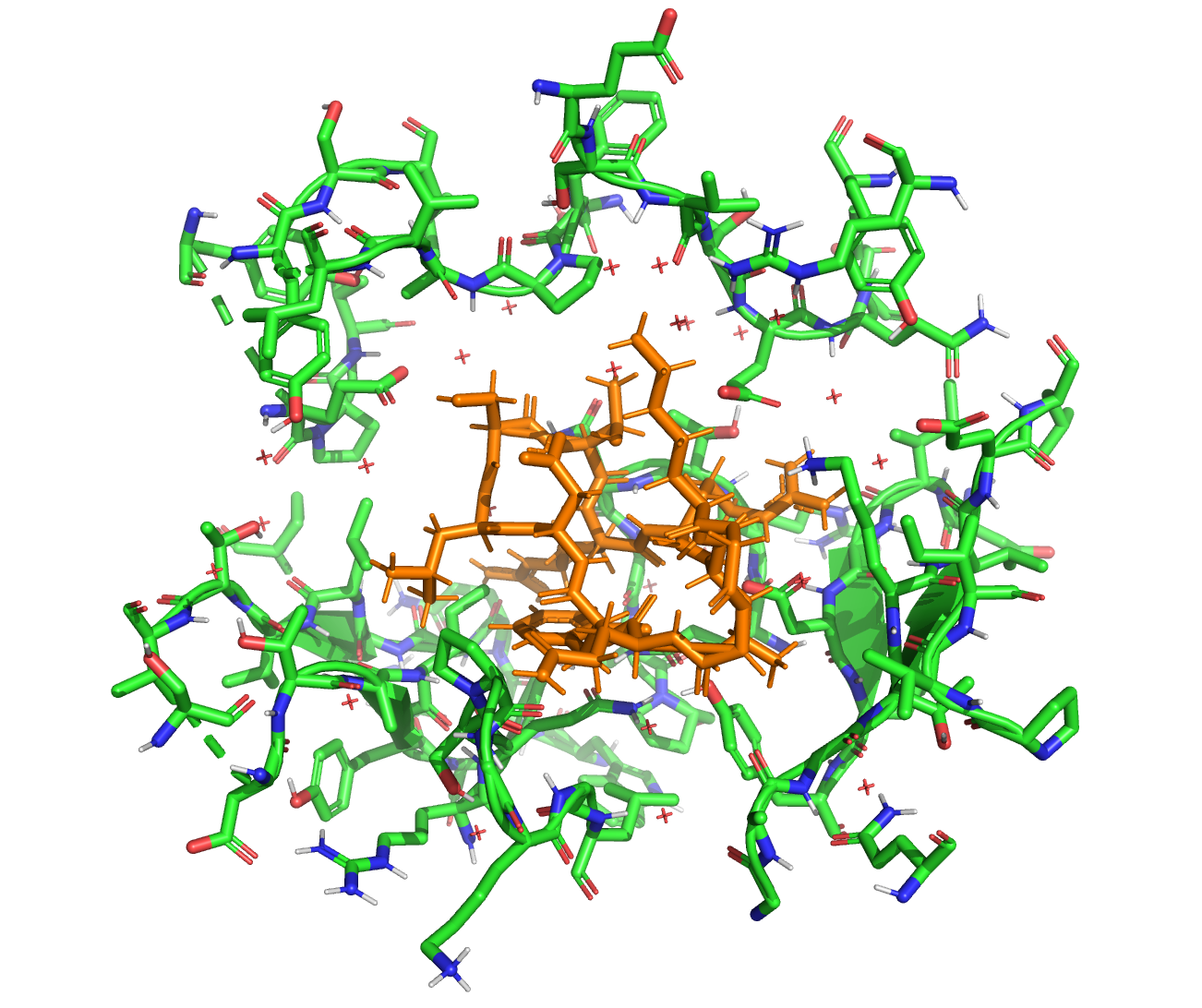} \\
    \end{tabular}
    \caption{%
    \textbf{Upper left}: t-SNE projections of ATOMICA embeddings of ligand–pocket interactions from PDBbind. The three colors show projected embeddings from three different test target clusters. One sample from each cluster is indicated by a star. The complexes corresponding to the three stars are shown in the remaining plots of this figure. These projections indicate that the ATOMICA space is organized by pocket shape, with some pockets forming tight regions. Considered together with Figure~\ref{fig:tsne_aff}, they suggest that these regions align with specific ranges of ligand sizes and affinities; in our experiments, scoring interactions representing such regions was particularly difficult when no representatives were available during training.
    \textbf{Upper right}: 3HKU, example from the 3DD0 cluster. 
    \textbf{Lower right}: 3QAA, example from the 3O9I cluster. 
    \textbf{Lower left}: 5ITF, example from the 3F3E cluster.
    }
    \label{fig:tsne_cluster}
\end{figure}

\begin{figure}[ht!]
  \centering
  \includegraphics[width=\textwidth]{figures/OOD_panel_suppl.png}
  \caption{Evolution of the scoring power of the three scoring methods GenScore, GEMS, and GEMS\textsubscript{ATOMICA} during model training, evaluated on CASF-2016 and the out-of-distribution (OOD) clusters 3DD0, 3F3E, 1NVQ, 3O9I and 1SQA. The x-axis shows training progress as a percentage of total training epochs, while the y-axis displays the Pearson correlation coefficient $\uparrow$ between predicted and true affinity. Each line represents the mean performance across five cross-validation folds, with shaded uncertainty regions ($\pm1$ standard deviation) indicating the variability in performance across the five training runs. Note that the uncertainty regions disappear towards higher epoch numbers due to early stopping, which results in different models completing training at different epochs. Consequently, the later portions of the curves are based on fewer than five models. 
  }
  \label{fig:OOD_training_suppl}
\end{figure}

\newpage
\begin{figure}[ht!]
  \centering
  \includegraphics[width=\textwidth]{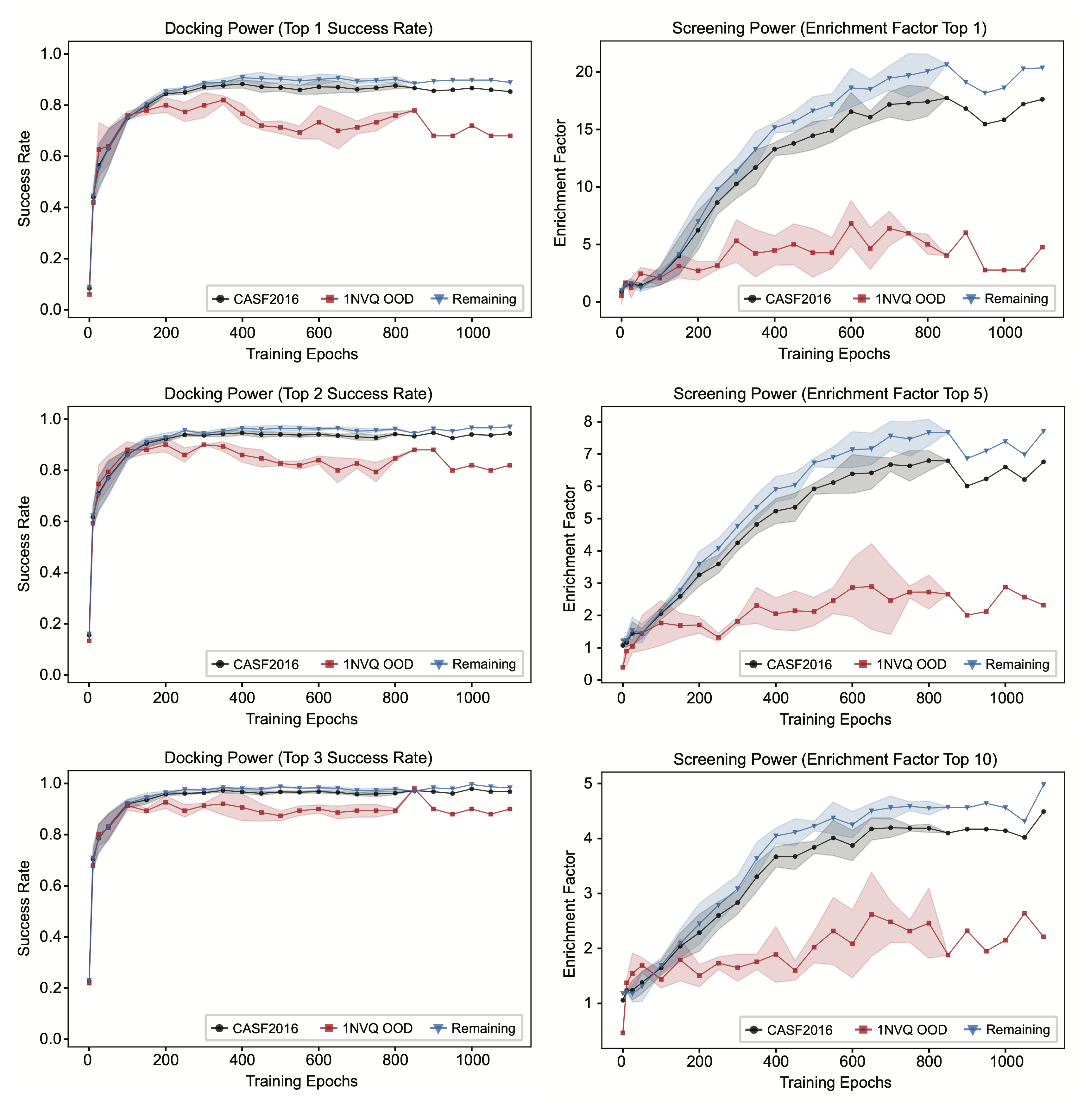}
  \caption{Evolution of docking power (left column) and screening power (right column) of GenScore on the complete decoy datasets of the CASF-2016 benchmark (black), the 1NVQ cluster subset (red), and the remaining decoy datasets with the 1NVQ cluster excluded (blue). The x-axis shows the absolute number of training epochs, while the y-axis displays the success rate (for docking) and the enrichment factors (for screening). Each line represents the mean performance across three cross-validation folds, with shaded regions indicating $\pm1$ standard deviation around the mean, providing a measure of variability in performance across the three independent model training runs. Note that the shaded uncertainty regions disappear towards the right end of each line due to early stopping, which results in different models completing training at different epochs. Consequently, the later portions of the curves are based on fewer than three models, precluding the calculation of meaningful standard deviations.}
  \label{fig:OOD_VS_training}
\end{figure}
\end{document}